\documentclass{article}
\PassOptionsToPackage{numbers,sort&compress,square}{natbib}

\usepackage[eandd, final]{neurips_2026}

\usepackage[dvipsnames]{xcolor}

\usepackage{xurl}
\usepackage{multicol, multirow, booktabs}
\usepackage{adjustbox}
\usepackage{appendix}
\usepackage{amsmath}
\usepackage{float}
\usepackage{graphicx}
\usepackage{colortbl}
\usepackage{soul}
\usepackage{pifont}
\usepackage{tikz}
\usepackage{algorithm}
\usepackage{algpseudocode}
\usepackage{caption}
\usepackage{xspace}

\usepackage{fontawesome5}
\usepackage{algpseudocode}
\usepackage{amsmath}
\usepackage[accsupp]{axessibility}
\usepackage[most]{tcolorbox}
\usepackage{enumitem} 

\newcommand{\bname}{SPOT-Bench}
\newcommand{\mname}{Ctrl-SPOT}

\newcommand{\cmark}{\textcolor{green!55!black}{\ding{51}}}
\newcommand{\xmark}{\textcolor{red!70!black}{\ding{55}}}

\definecolor{headergray}{RGB}{245,245,245}
\definecolor{headerblue}{RGB}{230,240,255}
\definecolor{headergreen}{RGB}{235,245,235}

\definecolor{headerdetection}{RGB}{250,242,236}
\definecolor{headerintervention}{RGB}{240,246,250}
\definecolor{headerinteraction}{RGB}{242,248,240}

\definecolor{promptbody}{RGB}{240,240,240}
\definecolor{promptheader}{RGB}{50,50,50}

\definecolor{cadmiumgreen}{rgb}{0.0, 0.42, 0.24}
\definecolor{darkpastelgreen}{rgb}{0.01, 0.75, 0.24}
\definecolor{brightpink}{rgb}{1.0, 0.0, 0.5}
\definecolor{keywordcolor}{RGB}{0, 102, 204}
\definecolor{commentcolor}{RGB}{0, 153, 0}
\definecolor{functioncolor}{RGB}{153, 0, 153}
\definecolor{cvprblue}{rgb}{0.21,0.49,0.74}

\makeatletter
\DeclareRobustCommand\onedot{\futurelet\@let@token\@onedot}
\def\@onedot{\ifx\@let@token.\else.\null\fi\xspace}
\def\eg{\emph{e.g}\onedot} 
\def\ie{\emph{i.e}\onedot}

\newtcolorbox[auto counter]{promptbox}[2][]{
    colback=promptbody,      
    colframe=promptheader,   
    fonttitle=\bfseries,     
    coltitle=white,          
    title={Prompt \thetcbcounter: #2},              
    label={#1},
    fontupper=\small\ttfamily, 
    arc=3mm,                 
    boxrule=1pt,             
    left=5pt, right=5pt, top=5pt, bottom=5pt 
}
\usepackage[utf8]{inputenc}
\usepackage[T1]{fontenc}
\usepackage{hyperref}
\usepackage{url}
\usepackage{booktabs}
\usepackage{amsfonts}
\usepackage{nicefrac}
\usepackage{microtype}
\usepackage{xcolor}
\usepackage[capitalize]{cleveref}
\usepackage{svg}

\crefname{figure}{Fig.}{Figs.}
\crefname{table}{Tab.}{Tabs.}
\crefname{algorithm}{Alg.}{Algs.}
\crefname{appendix}{App.}{Apps.}
\Crefname{appendix}{Appendix}{Appendices}

\title{Don't Pause! \\ Every prediction matters in a streaming video}

\author{
        Dibyadip Chatterjee$^{1}$, \, 
        Zhanzhong Pang$^{1}$, \,
        Fadime Sener, \,
        Yale Song$^{2}$, \,
        Angela Yao$^{1}$ \vspace{0.1cm} \\ 
        $^{1}$National University of Singapore \qquad $^{2}$Google Inc. \vspace{0.1cm} \\
        \faGlobe\ \href{https://dibschat.github.io/SPOT-Bench}{Page}
        \quad
        \faGithub\ \href{https://github.com/dibschat/SPOT-Bench}{Code}
        \quad
        \raisebox{-0.3em}{\includegraphics[height=1.4em]{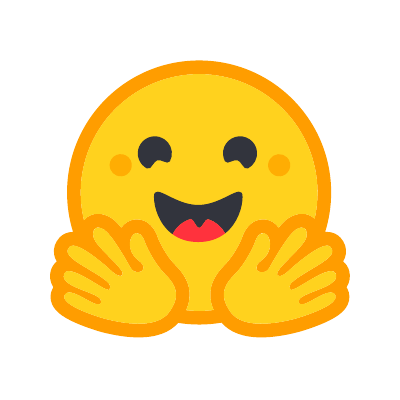}}\ \href{https://huggingface.co/datasets/cvml-nus/spot-bench}{Data}
}

\begin{document}

\maketitle

\begin{figure}[h]
    \centering
        \vspace{-0.5cm}
        \includegraphics[width=\linewidth]{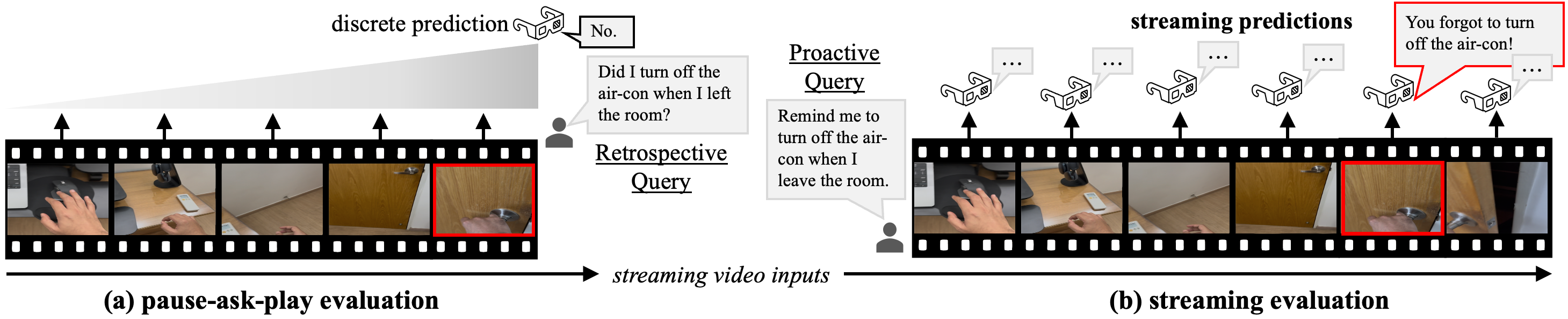}
        \vspace{-0.4cm}
        \captionof{figure}{
    \textbf{a) Pause-ask-play evaluation.} The stream is paused to pose a retrospective query; models may take streaming inputs but generate discrete outputs.~\textbf{(b) Streaming evaluation.} The model continuously monitors the live stream, deciding at every frame whether to respond or remain silent.}
    \label{fig:teaser}
\end{figure}

\vspace{0.2cm}
\begin{abstract}
\vspace{-0.3cm}
Streaming video models should respond at the moment an event unfolds, not after the moment has passed. Yet existing online VideoQA benchmarks remain largely retrospective. They pause the video at fixed timestamps, pose questions about current or past events, and score models only at those moments. This protocol leaves streaming outputs untested. To close this gap, we introduce SPOT-Bench, featuring multi-turn proactive queries that evaluate general streaming perception required by an always-on, real-time assistant. SPOT-Bench comes with Timeliness-F1, a consolidated metric that measures streaming outputs by their temporal precision and balanced coverage across the entire video. Our benchmark reveals that during streaming inference:~\textit{(i)}~MLLMs detect events reliably in a stream but spam predictions unprompted;~\textit{(ii)}~post-training MLLMs for silence reduces spamming but induces unresponsiveness;~\textit{(iii)} half of the streaming video expects no response, which we term dead-time --- compute spent here does not affect response latency. These findings motivate \mname{}, a training-free streaming controller for MLLMs, that retains their event perception while controlling their streaming outputs for balanced coverage. This establishes a strong baseline on \bname{}.
\vspace{-0.2cm}
\end{abstract}

\section{Introduction}\label{sec:intro}
\begin{figure}[t!]
    \centering
        \includegraphics[width=0.95\linewidth]{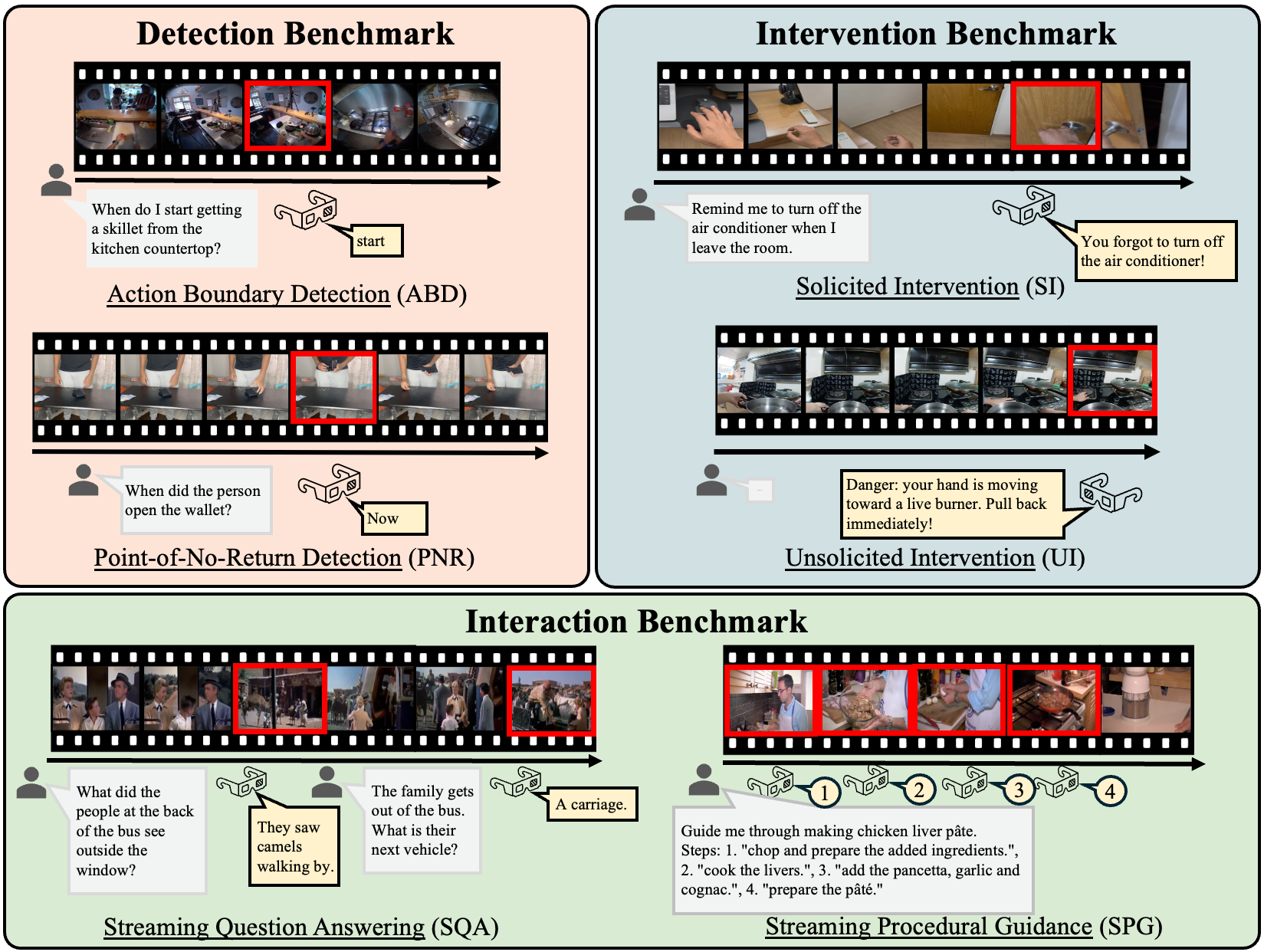}
        \vspace{-0.05cm}
        \captionof{figure}{
    \textbf{Overview of \bname{} tasks.} Each example illustrates proactive queries expecting streaming predictions over time with red boxes marking keyframes. 
    \label{fig:benchpreview}
    \vspace{-0.4cm}
}
\end{figure}

Consider a blind individual unknowingly reaching toward a live stove~(\cref{fig:benchpreview} UI), or an aircrew member overlooking a critical step during a pre-flight inspection~\cite{national1988aircraft}. In such fleeting moments of unawareness, a timely intervention could spare costly mistakes from disastrous consequences. As augmented reality (AR) devices and intelligent assistants~\cite{metaOrionGlasses, projectariaAriaFrom} become commonplace~\cite{10.1145/3749513, bellos2025towards, huh2025vid2coach}, these scenarios reveal new challenges beyond conventional video understanding~\cite{li2024llava, maaz2023video, mangalam2023egoschema, fu2025video}. Streaming Video Large Language Models (VideoLLMs) aim to address these demands by processing frames incrementally, as they arrive in a video stream, for online video question-answering (online VideoQA)~\cite{chen2024videollm, zhang2024flash, di2025streaming, qian2025dispider, yao2025timechat, wang2025omnimmi, chatterjee2025streaming, wang2025streambridge}.

Despite rapid progress in streaming VideoLLM research, their streaming~\textit{outputs} remain largely understudied --- when deployed on a live stream, such a model needs to continuously decide at every frame whether to respond or remain silent, and if the former, what to respond. Even though existing Online VideoQA benchmarks~\cite{lin2024streamingbench, niu2025ovo, huang2025online, wang2025omnimmi} have advanced streaming VideoLLMs, their evaluation protocol leaves streaming predictions untested. OVO-Bench~\cite{niu2025ovo} and StreamingBench~\cite{lin2024streamingbench}, the most widely used online benchmarks, adopt a ``pause-ask-play'' setup: the video is paused at a fixed timestamp, a query about ongoing or past events is posed, the model responds once, and the video resumes. Predictions occur \emph{only} at these pauses. Models are evaluated at these predefined timestamps rather than continuously across the full video. This provides no signal about streaming outputs and leaves delayed, missed, or over-responses unpenalized.

This limitation, however, runs deeper than the evaluation: existing tasks are predominantly \textit{retrospective}. Consider two queries (\cref{fig:teaser}): ``Did I turn off the air-con when I left the room?'' vs. ``Remind me to turn off the air-con when I leave the room.'' The former is \textit{retrospective}, looking back at what has already happened, while the latter is \textit{proactive}, requiring anticipating what is about to happen and responding at the right moment when the user reaches for the door knob. Consequently, existing benchmarks offer little incentive for such streaming proactive behavior~(\cref{ssec:incentivize}).

To address the limitations of pause-ask-play evaluation and retrospective task design, we introduce the~\textbf{S}treaming \textbf{P}erception \textbf{O}ver \textbf{T}ime \textbf{Bench}mark (\textbf{\bname{}}): a fully proactive, multi-turn, multi-response benchmark comprising 940 videos. \bname{} is built around six streaming tasks grouped into three categories~(\cref{fig:benchpreview}). \underline{Detection} tasks evaluate the fundamental streaming perception skills required for timely interactions. \underline{Interaction} and \underline{Intervention} tasks simulate realistic streaming settings, serving as diagnostic probes that expose the gap between current streaming VideoLLMs and the demands of real-world deployment. Evaluating these tasks requires measuring not just what a model predicts, but also \textit{when} it responds (and when it shouldn't). To this end, we introduce~\underline{Timeliness-F1}\footnote{Timeliness is a measure of how well a model responds at the right moments in the stream, not wall-clock latency.} (T-F1), a consolidated streaming evaluation metric that continuously assesses model responses along three axes: timeliness, multi-turn instruction following, and response frequency. Together, they reward well-timed predictions across the entire video, capture degradation across turns, and penalize over- and under-responses.

We benchmark a suite of closed- and open-source MLLMs on SPOT-Bench, evaluating their streaming outputs under proactive queries. Closed-source models~\cite{gpt5, pichai2025new} lead by a wide margin. Open-source models~\cite{zhang2025videollama, wang2025internvl3, cho2025perceptionlm, bai2025qwen2, bai2025qwen3} primarily collapse on Interaction and Intervention tasks, revealing a significant gap to real-world deployment. Open-source streaming VideoLLMs~\cite{xu2025streamingvlm, wang2025mmduet2}, despite larger temporal context, perform worse. Interestingly, we find that event perception is not the main bottleneck; offline MLLMs already detect fine-grained events reliably. The real challenge is over-responding~\ie~during streaming inference, offline MLLMs flood the prediction stream with repetitive responses. A natural fix is post-training on streaming data with a silence objective~\cite{wang2025mmduet2}. This reduces over-responses ($\downarrow$~false positives) but at the cost of making the model unresponsive ($\uparrow$~false negatives). This tradeoff is invisible under existing pause-ask-play protocols and only surfaces with our T-F1.

Since MLLMs already detect events reliably, we ask: can training-free adaptation suffice for streaming inference? To this end, we introduce~\textbf{Ctrl-SPOT}, a training-free streaming output~\textbf{C}on\textbf{tr}o\textbf{l}ler that explicitly tracks query-response state across frames to suppress repetitive outputs without retraining. Continuous evaluation with T-F1 further reveals a structural property of streaming inference: between query windows, most frames carry no active query, resulting in substantial \textit{dead-time}. \mname{} exploits this idle period for periodic summarization and memory updates, extending temporal context without increasing response latency. Being training-free, \mname{} preserves the event perception of its backbone, thus establishing a strong reference baseline on~\bname{} while transferring effectively to retrospective benchmarks~\cite{lin2024streamingbench, niu2025ovo}, where it achieves state-of-the-art performance.

To summarize, we (1) introduce \bname{}, a benchmark for evaluating proactive streaming behavior across six tasks, (2) propose Timeliness-F1 to measure how well a model responds at the right moments across the full video, and (3) introduce Ctrl-SPOT, a training-free streaming controller that enables strong proactive and retrospective streaming video understanding.

\section{Related Work}
\noindent\textbf{Streaming Video Large Language Models} have recently gained traction in Multimodal LLM research. Originally built for long video understanding under bounded memory and compute~\cite{he2024ma, song2024moviechat, azad2025hierarq, di2025streaming, yang2025streammem, chatterjee2025streaming}, they have since moved toward real-time understanding.~\cite{chen2024videollm} initiated this line of work, where a system monitors a live video and responds when needed while staying silent otherwise. Streaming VideoLLMs incrementally process frames at a predefined rate and aim to respond at critical moments~\cite{chen2024videollm, wu2024videollm, qian2025dispider, yao2025timechat}, sometimes proactively~\cite{qian2025dispider, wang2025streambridge, wang2025mmduet2, yang2026mage, yao2026joyai} and target low latency~\cite{chatterjee2025streaming, xu2025streamingvlm, pang2026decouple}. Despite this progress in processing streaming inputs, there has been limited research on handling streaming outputs and their evaluation. As a result, delayed, missed, or over-responses go unpenalized, making them impractical for always-on, real-time assistants.

\noindent\textbf{Online Video Understanding} studies how models make decisions over streaming content under partial observations. Traditionally, Online action detection (OAD)~\cite{de2016online,xu2021long} embodies this setting: models predict actions every frame as they unfold, without access to the future. Most OAD work focuses on action spotting in short videos, with recent extensions to longer procedural videos~\cite{song2023ego4d, pang2025context, kang2025open}. Action anticipation~\cite{pei2011parsing, lan2014hierarchical, damen2018scaling} has been another widely explored online video task~\cite{sener2020temporal, girdhar2021anticipative, gong2022future, lai2024human}. Only recently, to support VideoLLM research, several online VideoQA benchmarks have emerged~\cite{huang2025online, niu2025ovo, qian2024streaming, wang2025omnimmi, wang2025proactivevideoqa}; OVO-Bench~\cite{niu2025ovo} and StreamingBench~\cite{lin2024streamingbench} being the most widely used. Yet their tasks remain largely retrospective: models are queried about events that already occurred or about what is occurring. Consequently, in such benchmarks, most tasks can be solved from the query frame or its immediate past, without requiring per-frame streaming. Recent proactive benchmarks~\cite{wang2025proactivevideoqa} have attempted to move toward anticipative queries but fail to evaluate timeliness of the model across a full video. This breaks from the continuous evaluation paradigm established in OAD and leaves the notion of timeliness incomplete. Our \bname{} closes this gap by evaluating every streaming prediction and aligning the protocol with the demands of a real-time assistant.

\vspace{-0.2cm}
\section{Preliminaries: Online Video Question-Answering}\label{sec:prelims}
Here, we formally introduce online video question-answering and define two paradigms: retrospective VideoQA and proactive VideoQA~(\cref{fig:teaser}). Let $V=\{v_t\}_{t=1}^{T}$ be a video stream, where at any time $\tau$, the observed video is $V_{\le \tau}=\{v_t\}_{t=1}^{\tau}$. A streaming model $\mathcal{M}$ produces responses $R_\tau = \{r_{t_1}, r_{t_2},\cdots\}$ to a query $q_\tau$ posed at time $\tau$. A~\textit{turn} denotes a query paired with its response set ($q_\tau, R_\tau$). A \textit{multi-turn} interaction is a sequence of turns $\{(q_{\tau_i}, R_{\tau_i})\}_{i=1}^{N}$ with $\tau_1 < \tau_2 < \cdots < \tau_N$. In online VideoQA, for any turn, all responses must be strictly causal, satisfying $t_i \ge \tau$. \\

\noindent\textbf{Retrospective VideoQA.}
In this setting, all evidence needed to answer a query lies in the observed past. A query $q_\tau$ refers to events up to time $\tau$, and the model produces a response using only context $V_{\le \tau}$.
Queries are evaluated with a ``pause-ask-play'' protocol: the video is paused at time $\tau$, the query is posed, the model responds, and playback resumes. A video may contain multiple turns, but each yields exactly one response. In practice~\cite{lin2024streamingbench, niu2025ovo}, long videos are segmented into short clips, with the multi-turn structure represented as
\{\texttt{$<$}$V_{\le \tau_1}$\texttt{$>$}\texttt{$<$}$q_{\tau_1}$\texttt{$>$}\texttt{$<$}$r_{\tau_1}$\texttt{$>$},\;
\texttt{$<$}$V_{\le \tau_2}$\texttt{$>$}\texttt{$<$}$q_{\tau_2}$\texttt{$>$}\texttt{$<$}$r_{\tau_2}$\texttt{$>$}, $\cdots$\}. Evaluation therefore occurs only at query points $\tau_i$ and solely on response correctness.

\noindent\textbf{Proactive VideoQA.}
In this setting, the evidence required to answer a query lies in the future. A query $q_\tau$ initiates a causal monitoring process over an evolving video. The model observes an incrementally expanding visual context $V_{\le t}$ with $t>\tau$, whose endpoint is unknown at query time. The model may therefore emit one or more responses $r_t = \mathcal{M}(q_\tau, V_{\le t})$, whenever the visual evidence relevant to $q_\tau$ emerges, producing a temporally ordered set $R_\tau=\{r_{t_1}, r_{t_2}, \ldots\}$ with $\tau < t_1 \le t_2 \le \cdots$. Otherwise, the model remains silent. A video may contain one or more turns, and each turn can yield multiple responses over time ($|R_\tau|\geq1$). 
The resulting query-response structure can be represented as
\{\texttt{$<$}$v_1$\texttt{$>$}\,\texttt{$<$}$v_2$\texttt{$>$}\,\texttt{$<$}$v_3$\texttt{$>$}\,\texttt{$<$}$q_{3}$\texttt{$>$}\,\texttt{$<$}$v_4$\texttt{$>$}\,\texttt{$<$}$v_5$\texttt{$>$}\,\texttt{$<$}$r_{5}$\texttt{$>$}\,
\texttt{$<$}$v_6$\texttt{$>$}\,\texttt{$<$}$v_7$\texttt{$>$}\,\texttt{$<$}$v_8$\texttt{$>$}\,\texttt{$<$}$r_{8}$\texttt{$>$} $\cdots$\},
where $q_3$ triggers the responses $R_3=\{r_5, r_8\}$. Alternatively, an initial prompt $q_0$ may specify the task without further queries, in which case the model autonomously issues responses based on $V_{\le t}$~\ie~\{\texttt{$<$}$q_0$\texttt{$>$} \texttt{$<$}$v_1$\texttt{$>$} \texttt{$<$}$v_2$\texttt{$>$}
\texttt{$<$}$r_{2}$\texttt{$>$} \texttt{$<$}$v_3$\texttt{$>$}
\texttt{$<$}$v_4$\texttt{$>$} \texttt{$<$}$r_{4}$\texttt{$>$} $\cdots$\}.
Evaluation therefore must consider not only response correctness but also \textit{timeliness}, penalizing responses that are mistimed.
\vspace{-0.2cm}

\begin{table}[t]
\centering
\setlength{\tabcolsep}{3mm}{
\caption{\textbf{Pause-ask-play evaluation on existing Online (retrospective) VideoQA benchmarks.}} 
\label{tab:prelims:retroqa}
\resizebox{\linewidth}{!}{
\begin{tabular}{@{}l c c ccc ccc@{}}
\toprule
\textbf{Model} & \textbf{\#Frames} & \textbf{Model} &
\multicolumn{3}{c}{\textbf{OVO-Bench}~\cite{niu2025ovo}} &
\multicolumn{3}{c}{\textbf{StreamingBench}~\cite{lin2024streamingbench}} \\
\cmidrule(lr){4-6}\cmidrule(lr){7-9}
 & & \textbf{Size} & \textbf{Real-Time} & \textbf{Backward} & \textbf{Forward} & \textbf{Real-Time} & \textbf{Omni} & \textbf{Contextual} \\
\midrule

\;\; Chance Performance & & & 25.0 & 26.8 & 33.3 & 25.0 & 25.0 & 18.8 \\
\rowcolor{headergray}
\multicolumn{9}{l}{\textbf{Blind Baseline}} \\
\quad GPT-5~\cite{gpt5} & 0 & & 36.6 & 58.1 & 32.3 & 58.2 & 36.6 & 27.8 \\
\quad Gemini-3-Flash~\cite{pichai2025new} & 0 & & 44.8 & 55.3 & 33.9 & 59.8 & 34.8 & 29.0 \\
\quad Qwen2.5-VL~\cite{bai2025qwen2} & 0 & 7B & 30.0 & 49.1 & 28.7 & 52.5 & 32.7 & 21.8 \\
\quad Qwen3-VL~\cite{bai2025qwen3} & 0 & 8B & 29.2 & 44.3 & 32.0 & 48.7 & 32.5 & 24.6 \\

\rowcolor{headergray}
\multicolumn{9}{l}{\textbf{Single (query) frame Baseline}} \\
\quad GPT-5~\cite{gpt5} & 1 & & 70.4 & 51.8 & 40.5 & 74.0 & 52.7 & 56.9 \\
\quad Gemini-3-Flash~\cite{pichai2025new} & 1 & & 81.8 & 64.4 & 45.2 & 82.9 & 57.1 & 68.1 \\
\quad Qwen2.5-VL~\cite{bai2025qwen2} & 1 & 7B & 67.8 & 44.0 & 36.9 & 75.5 & 48.4 & 53.6 \\
\quad Qwen3-VL~\cite{bai2025qwen3} & 1 & 8B & 74.7 & 50.3 & 40.8 & 76.0 & 52.1 & 59.2 \\
\midrule

\rowcolor{headergray}
\multicolumn{9}{l}{\textbf{Recency baseline (4 recent frames at 1fps)}} \\
\quad Qwen2.5-VL~\cite{bai2025qwen2} & 4 & 7B & 75.3 & 42.3 & 39.0 & 79.0 & 64.6 & 56.9 \\
\quad Qwen3-VL~\cite{bai2025qwen3} & 4 & 8B & \textbf{82.3} & 49.3 & 45.0 & \textbf{79.6} & \textbf{67.2} & \textbf{62.9} \\

\rowcolor{headergreen}
\multicolumn{9}{l}{\textbf{State-of-the-art (open-source) streaming VideoLLMs}} \\
\quad Timechat-Online~\cite{yao2025timechat} & 1fps & 7B & 61.9 & 41.7 & 36.7 & 75.4 & 37.8 & 35.3 \\
\quad StreamAgent~\cite{yang2025streamagent} & 1fps & 7B & 61.3 & 41.7 & 45.4 & 74.3 & 36.3 & 34.6 \\
\quad StreamBridge~\cite{wang2025streambridge} & 1fps & 7B & 71.3 & \textbf{68.0} & 48.4 & 77.0 & 24.1 & 32.6 \\
\quad StreamForest~\cite{zeng2025streamforest} & 1fps & 7B & 61.2 & 52.0 & \textbf{53.5} & 77.3 & - & - \\
\bottomrule
\end{tabular}
}}
\vspace{-0.4cm}
\end{table}

\subsection{Do existing Online VideoQA benchmarks incentivize streaming models?}\label{ssec:incentivize}
\vspace{-0.2cm}
The most widely used Online VideoQA benchmarks, StreamingBench~\cite{lin2024streamingbench} and OVO-Bench~\cite{niu2025ovo} have become the primary testbeds for streaming model development~\cite{yao2025timechat, wang2025streambridge, zeng2025streamforest, yang2025streamagent}. Both benchmarks adopt retrospective queries under a ``pause-ask-play'' evaluation protocol. Here, we examine whether they incentivize streaming model development~(\cref{tab:prelims:retroqa}). To this end, we compare state-of-the-art streaming models with closed- and open-source offline MLLMs and observe the following:

\noindent\textbf{Blind baselines beat chance.} These are multiple-choice benchmarks, and strong LLMs (GPT, Gemini) can exploit structural cues in the QAs to guess correctly without any visual input. \\
\noindent\textbf{One frame is nearly enough.}
A single query frame surpasses streaming models on Real-Time tasks, and remains competitive across the board, including on StreamingBench's proactive task (GPT-5: 58\%, Gemini: 55\%, Qwen2.5-VL: 39\%, Qwen3-VL: 47\%). \\
\noindent\textbf{Four frames outperform streaming models.} The only exception is OVO-Bench's Backward split, which by definition benefits from longer temporal context. Surprisingly, even on StreamingBench's Omni-Source tasks, which require audio-visual understanding, two-thirds of queries can be answered correctly using only 4 frames, without audio.

\noindent\textbf{Conclusion.}
Together, these findings reveal a fundamental limitation: OVO-Bench and StreamingBench measure descriptive understanding, evaluating whether a model can describe observed events. Their evaluation provides no signal for timeliness or frequency. Their short durations remove the need for long-term temporal modeling --- an offline model processing a single frame can match one that streams every frame. Consequently, these benchmarks offer little incentive to develop streaming models with proactive, timely intervention capabilities. \bname{} is designed to address this gap.

\section{\bname{}}\label{sec:spot_bench}
SPOT-Bench is designed with the following goals: 
(1)~queries target ongoing and future events rather than past ones, forcing the model to predict under incomplete or partial observations;
(2)~streaming model outputs must demonstrate timeliness; and 
(3) evaluation should be continuous across the entire video such that every prediction contributes to the final score, positively or negatively.
We source videos from existing datasets~(\cref{tab:task_stats}) and follow a semi-automatic, human-in-the-loop annotation pipeline. To ensure annotation quality, all final annotations were verified by two independent human reviewers. We provide the full data collection and annotation pipeline in~\cref{supp::ssec:annotation}.

\begin{table}[t]
\centering
\setlength{\tabcolsep}{0.5mm}{
\caption{\textbf{Comparison of Online VideoQA benchmarks.} sorted in ascending order of \%Proactive. 
\textbf{Concurrent}QA, \textbf{Unsolicited} and \textbf{Streaming Evaluation} is unique to \bname{}.}
\label{tab:bench_compare}
\resizebox{0.99\linewidth}{!}{
\begin{tabular}{@{}lcccccccccc@{}}
\toprule
\textbf{Benchmark} & \textbf{\#Videos} & \textbf{\#Slots} & \textbf{\%Proactive} & \textbf{Multi-turn} & \textbf{Multi-response} & \textbf{Referential} & \textbf{Concurrent} & \textbf{Unsolicited} & \textbf{Stream. Eval} \\
\midrule
OV-Bench~\cite{huang2025online}                    & 1463 & 7090 & 0\%   & \cmark & \xmark & \cmark & \xmark & \xmark & \xmark \\
StreamBench~\cite{xiong2025streaming}              & 275  & 1838 & 0\%   & \cmark & \xmark & \xmark & \xmark & \xmark & \xmark \\
StreamingBench~\cite{lin2024streamingbench}        & 900  & 4500 & 5.5\% & \cmark & \xmark & \xmark & \xmark & \xmark & \xmark \\
OVO-Bench~\cite{niu2025ovo}                        & 644  & 1640 & 10.5\% & \cmark & \cmark & \xmark & \xmark & \xmark & \xmark \\
OmniMMI~\cite{wang2025omnimmi}                     & 1121 & 2290 & 22\%  & \cmark & \xmark & \xmark & \xmark & \xmark & \xmark \\
VideoLLM-online~\cite{chen2024videollm}            & 1165 & 1165 & 100\% & \xmark & \cmark & \xmark & \xmark & \xmark & \xmark \\
ProactiveVideoQA~\cite{wang2025proactivevideoqa}   & 1377 & 1427 & 100\% & \xmark & \cmark & \xmark & \xmark & \xmark & \xmark \\
\midrule
\textbf{\bname{} (ours)}                           & 940 & 5093 & 100\% & \cmark & \cmark & \cmark & \cmark & \cmark & \cmark \\
\bottomrule
\end{tabular}}}
\vspace{-0.3cm}
\end{table}

\begin{table}[t]
\centering
\caption{\textbf{\bname{} Task Taxonomy.} We report dataset scale (\#Slots refers to response slots) and per-video statistics reflecting streaming density for each task in~\bname{}. \textbf{MT/MR} indicates multi-turn and multi-response respectively. \textbf{Inter-annotator agreement} for human preference intervals is reported as mean absolute deviation (MAD) in secs. 
The interquartile range (IQR) in seconds determines the task-specific timeliness intervals ($t_e-t_s=2\cdot\mathrm{IQR}$).}
\label{tab:task_stats}
\setlength{\tabcolsep}{1mm}{
\resizebox{\linewidth}{!}{
\begin{tabular}{@{}l c c c c c c c c l@{}}
\toprule
& & & & \multicolumn{3}{c}{\textbf{Avg. per Video}} 
& \multicolumn{2}{c}{\textbf{Inter-anno agreement}} 
& \\
\cmidrule(lr){5-7} \cmidrule(lr){8-9}
\textbf{Task} 
& \textbf{Acr.}
& \textbf{\#Slots} 
& \textbf{MT/MR} 
& \textbf{Turns} 
& \textbf{Resp.} 
& \textbf{Dur. (s)}
& \multicolumn{1}{c}{\textbf{MAD (s)}}
& \multicolumn{1}{c}{\textbf{IQR (s)}}
& \textbf{Source data} \\
\midrule

\rowcolor{headerdetection}
\multicolumn{10}{l}{\textbf{Detection}} \\
Action Boundary Detection & ABD & 2340 & \cmark/\cmark & 5.0 & 1.4 & 200.7 & 0.4 & 1.0 & EgoExo4D~\cite{grauman2024ego}, HT-Step~\cite{afouras2023ht} \\
Point-of-No-Return Detection & PNR & 1234 & \cmark/\cmark & 4.1 & 1.1 & 242.1 & 0.2 & 0.5 & Ego4D~\cite{grauman2022ego4d}, THUMOS'15~\cite{yeung2018every}, PT~\cite{patraucean2023perception}\\
\addlinespace[0.6ex]

\rowcolor{headerinteraction}
\multicolumn{10}{l}{\textbf{Interaction}} \\
Streaming Question Answering & SQA & 819 & \cmark/\xmark & 8.6 & 1.0 & 519.1 & 1.1 & 2.5 & MovieNet~\cite{huang2020movienet}, Ego4D, EgoExo4D, HT-Step \\
Streaming Procedure Guidance & SPG & 322 & \xmark/\cmark & 1.0 & 7.0 & 354.9 & 0.6 & 1.5 & Ego4D Goal-Step~\cite{song2023ego4d}, HT-Step \\
\addlinespace[0.6ex]

\rowcolor{headerintervention}
\multicolumn{10}{l}{\textbf{Intervention}} \\
Solicited Intervention & SI & 252 & \xmark/\cmark & 1.0 & 2.6 & 277.0 & 0.5 & 1.0 & HoloAssist~\cite{wang2023holoassist} \\
Unsolicited Intervention & UI & 126 & \xmark/\cmark & 1.0 & 1.3 & 40.4 & 0.3 & 0.5 & EgoBlind~\cite{xiao2025egoblind} \\

\bottomrule
\end{tabular}}}
\vspace{-0.25cm}
\end{table}

\subsection{Benchmark Tasks}\label{sec:ssec::tasks}
In accordance with our objectives, we introduce six streaming tasks grouped into three broad categories~(\cref{tab:task_stats}). Refer to~\cref{fig:benchpreview} for illustrated examples of each task.

\noindent\textbf{1) Detection} is designed to evaluate a model’s online event perception in a streaming setting~\ie~can it identify the exact moment an event occurs, given only partial observations up to that point?

\noindent\textbf{i) Action Boundary Detection (ABD).} 
This task reformulates online action detection as a \textit{point-wise} variant conditioned on a query. Unlike OAD, which performs dense frame-level classification~\cite{de2016online, zhong2024onlinetas}, ABD requires identifying the temporal boundary points $(\bar{s}, \bar{e})$ of actions as they unfold in real time. Given a streaming video $V$ and a boundary query $q_\tau$ ($\tau<\bar{s}<\bar{e}$), the model must respond only at the timestamps (${\hat{s}, \hat{e}}$) that align temporally with the ground-truth boundaries (${\bar{s}, \bar{e}}$). \\
\noindent\textbf{ii) Point-of-No-Return Detection (PNR).}
PNR is defined in Ego4D~\cite{grauman2022ego4d} as the onset of an irreversible object state change. In our setting, we reinterpret this idea as identifying a single critical temporal point within an action segment. Given $V_{\le t}$, the model must respond only at the PNR point(s) that align with the ground truth. Multiple PNR points may exist within a single video for a given query.

\noindent\textbf{2) Interaction} evaluates a model’s ability to engage in streaming interactions, producing temporally grounded, context-aware responses.

\noindent\textbf{iii) Streaming Question Answering (SQA).}
SQA represents a realistic interactive setting where users often ask questions slightly before an event occurs, requiring the model to anticipate future states and respond under partial observation. Given $V_{\le t}$ and $q_\tau$, the model must decide when and what to answer at a future timestamp once sufficient visual evidence is available. \\
\noindent\textbf{iv) Streaming Procedure Guidance (SPG).}
This is a proactive responding task where the model assists a user performing some procedural activity (\eg~cooking) in real-time.
Given a context prompt $q_0$ describing the task, its step sequence, and a streaming video $V_{\le t}$, the model must continuously monitor user progress and issue guidance responses $R_0$ that instruct the next step when the current one is nearing completion. This requires recognizing the current step, estimating progress, and anticipating the next, all at the same time, making SPG particularly challenging in real-world scenarios.

\noindent\textbf{3) Intervention} evaluates scenarios where the model assists a user without explicit requests, purely by anticipating potential errors or failures

\noindent\textbf{v) Solicited Intervention (SI).}
Here, the model assists a user executing a known procedure but intervenes only when the user hesitates or makes an error. Given $V$ and a context prompt $q_0$ describing the procedure, the model must monitor user performance and decide when and what corrective response(s) $R$ to issue as the execution deviates from the expected sequence. This task is highly challenging, as the model must continuously anticipate future states, align them with the current context, and determine precisely when to intervene before it is too late. \\
\noindent\textbf{vi) Unsolicited Intervention (UI).}
This task represents the highest level of autonomy in SPOT-Bench. It focuses on safety-critical scenarios where intervention becomes an objective necessity rather than a subjective choice. Unlike SI, there is no prior task context ($q_0$)~\ie~the user operates freely without any predefined structure. The model must recognize imminent risk and intervene at the critical moment, early enough to prevent failure, yet restrained enough to avoid unnecessary or disruptive responses.

\vspace{-0.3cm}
\subsection{Streaming Evaluation Metric}\label{sec:ssec::evaluation}
Prior works in streaming detection~\cite{de2016online, shou2018online, pang2025context, nowozin2012action} treat actions as discrete~\textit{points} in time, focusing on the exact moment an event starts or ends. However, such pointwise formulations fail to capture the temporal ambiguity of human perception. In the SQA query ``A policeman is driving his car, what is his destination?'', the correct response is not tied to a single instant but spans a short interval during which sufficient visual evidence emerges. Similarly, for the ABD query ``When does the camera wearer start inflating the tire with an air pump?'', even human annotators disagree on the onset of the action. Such cases highlight that human perception of actions is~\textit{interval-based}, not instantaneous.

\noindent\textbf{Calibrating to human perception intervals.}
To capture this temporal ambiguity, we conduct a controlled human study. We recruit over 20 annotators, of which 10 meet our quality bar and annotate acceptable response intervals across representative samples for each task. After filtering out outliers, eight annotations per task having the least intra-task variance were retained to establish our human preference intervals. We observe that annotation agreement correlates with task granularity. The resulting intervals are: $\pm$2.5 s for SQA, $\pm$1.5 s for SPG, $\pm$1 s for SI and ABD, and $\pm$0.5 s for UI and PNR. These \textit{gold} intervals define the temporal range of annotator agreement, which we adopt as the ground truth for response timeliness. Additional details are provided in~\cref{supp:ssec:human-study}.

\begin{figure}[t]
\centering
\vspace{0.3cm}
\includegraphics[width=\linewidth]{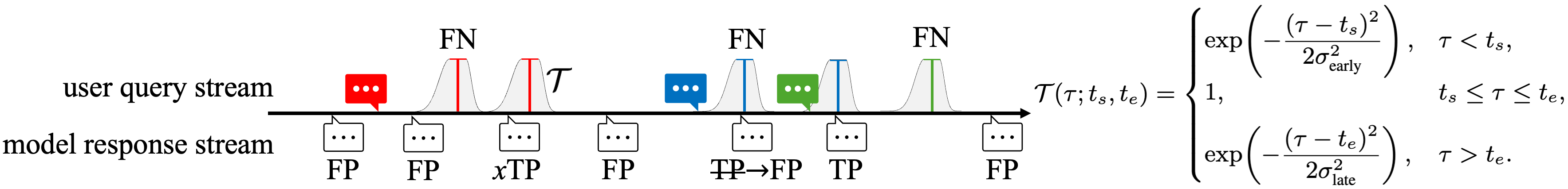}
\caption{\textbf{Streaming Evaluation Protocol}. The greedy matching procedure computes true positives (TP), false positives (FP), and false negatives (FN): TPs are temporally valid and semantically matched predictions; FPs are unmatched; FNs are unfilled slots. \textbf{$\mathcal{T}$-score} (right) penalizes mistimed predictions asymmetrically, favoring early responses over late ones.
}
\vspace{-0.5cm}
\label{fig:evaluation}
\end{figure}

\noindent\textbf{Timeliness Score.}
While gold intervals capture human consensus, they overlook the causal nature of streaming. Early responses are often preferable for enabling intervention or guidance before an event unfolds, whereas delayed responses may still be useful in VideoQA when additional context improves accuracy. To reflect this, we define a Timeliness-score ($\mathcal{T}$-score) that penalizes both early and late predictions, but with a stricter penalty for late ones, favoring anticipative behavior over reactive delay. To model this asymmetry, we define a weighting function centered on the human-calibrated gold interval $[t_s, t_e]$. Within this interval, predictions are considered fully timely and receive a score of 1. Beyond it, the score decays smoothly with Gaussian tails, using a slower decay for early responses and a sharper one for late responses~\cref{fig:evaluation}. Here, $\sigma_\text{early} > \sigma_\text{late}$ controls the asymmetric variance, producing a wider, slower decay for early predictions and a narrower, steeper falloff for late ones.

\noindent\textbf{Timeliness-F1.} 
\textit{Every prediction matters in a streaming video.} Timeliness-F1 captures exactly that, ``How well a model responds across the entire video horizon?'', rewarding timely, accurate predictions while penalizing both over- and under-responses. Each gold interval $[t_s, t_e]$ serves as a \textit{slot}, and model responses are evaluated causally against these slots. For each prediction, we compute its $\mathcal{T}$-score~(\cref{fig:evaluation}) with respect to every unfilled slot. Consequently, predictions with non-zero $\mathcal{T}$-score are semantically checked against the ground-truth response. A prediction is valid only if both $\mathcal{T}>0$ and it passes the semantic check. A valid response cannot be well-timed but semantically wrong.

Among all eligible slots, each slot is greedily matched to the prediction with the highest $\mathcal{T}$, i.e., the closest slot in time. Once filled, a slot is closed; subsequent matches are treated as false positives (FPs). Unmatched predictions are also FPs, while unfilled slots are false negatives (FNs). For tasks with multiple valid responses, each gold interval is treated as an independent slot, filled with the same greedy rule. Matched predictions receive fractional credit based on their $\mathcal{T}$ score: full credit is awarded if the prediction falls within the gold interval, partial credit (\textit{x}TP) otherwise~(\cref{fig:evaluation}). Precision and recall are computed using the sum of these weighted TPs, and their harmonic mean defines our~\underline{Timeliness-F1}. The metric jointly reflects response accuracy, timeliness, and frequency: low precision indicates over-response, while low recall indicates under-response. In practice, we use Timeliness-F1@K. The full algorithm is provided in~\cref{alg:supp::greedy_matching} with additional details in~\cref{supp::ssec:t-f1}.

\section{Online Experiments}\label{ssec:online_bench}
General-purpose MLLMs are natural backbone candidates for streaming models. In this section, we estimate their online event perception abilities in isolation~\ie~how timely a model detects an event. 
These models take a batch of frames and produce a single response. They cannot ingest streaming frames; as a result, running them afresh at every frame is prohibitively expensive. Also, they are not trained to control their output, or decide when to stay silent. As a result, to measure their event perception fairly under a controlled protocol, we simulate a streaming first-match sweep:
(i) Each video is segmented by its GT responses and their gold intervals ($\mathcal{T}>0$) comprise a window.
(ii) For every second in this window, the model is evaluated using the preceding one-minute visual context (uniformly sampled into 64 frames).
(iii) The sweep exits early when a match is found.
This controlled protocol isolates event perception from streaming behavior, ensuring models are evaluated on what events they can perceive from incoming (partial) observations.

\noindent\textbf{Baselines.}
We evaluate the strongest closed-source MLLMs, \textbf{GPT-5}~\cite{gpt5} and \textbf{Gemini-3-Flash}~\cite{pichai2025new}, to approximate the perception ceiling. We also evaluate leading open-source MLLMs: \textbf{VideoLLaMA3}~\cite{zhang2025videollama}, \textbf{InternVL-3.5}~\cite{wang2025internvl3}, \textbf{PerceptionLM}~\cite{cho2025perceptionlm}, \textbf{Qwen2.5-VL}~\cite{bai2025qwen2}, and \textbf{Qwen3-VL}~\cite{bai2025qwen3}.

\noindent\textbf{Evaluation Metrics.}\label{ssec:online_metrics} $\mathcal{T}$-score@1 is the natural choice to quantify first-match sweep. Since our simulated protocol does not run over the full video, FPs are artificially suppressed, rendering T-F1 uninformative. The closer the first match is to the gold interval, the higher the $\mathcal{T}$-score@1. 

\begin{table}[t]
\centering
\setlength{\tabcolsep}{4mm}{
\caption{\textbf{Online evaluation MLLMs' event perception on~\bname{}} using $\mathcal{T}$-score@1 ($\uparrow$).}
\label{tab:offlineevals}
\resizebox{\linewidth}{!}{
\begin{tabular}{@{}l c c cc cc cc c@{}}
\toprule
\textbf{Model} & \textbf{\#Frames} & \textbf{Model} &
\multicolumn{2}{c}{\textbf{Detection}} &
\multicolumn{2}{c}{\textbf{Interaction}} &
\multicolumn{2}{c}{\textbf{Intervention}} &
\textbf{Overall} \\
\cmidrule(lr){4-5}\cmidrule(lr){6-7}\cmidrule(lr){8-9}
& & \textbf{Size} & \textbf{ABD} & \textbf{PNR} & \textbf{SQA} & \textbf{SPG} & \textbf{SI} & \textbf{UI} & \textbf{Avg.}\\
\midrule
\rowcolor{headergray}
\multicolumn{10}{l}{\textbf{Closed-source MLLMs}} \\
GPT-5~\cite{gpt5} (Blind baseline) & 0 & & 0.4 & 0.3 & 0.0 & 0.0 & 0.0 & 0.0 & 0.1 \\
GPT-5~\cite{gpt5} & 64 & & \textbf{55.6} & \textbf{36.1} & \textbf{46.2} & \textbf{30.9} & 28.4 & \textbf{55.4} & \textbf{42.1} \\
Gemini-3-Flash~\cite{pichai2025new} & 64 & & 46.6 & 30.1 & 44.1 & 29.1 & \textbf{30.2} & 22.0 & 33.7 \\
\addlinespace[0.6ex]
\rowcolor{headergray}
\multicolumn{10}{l}{\textbf{Open-source MLLMs}} \\
VideoLLaMA3~\cite{zhang2025videollama} & 64 & 7B & 40.2 & 12.4 & 16.0 & 0.0 & 0.0 & 0.0 & 11.4 \\
InternVL-3.5~\cite{wang2025internvl3} & 64 & 8B & 41.6 & \underline{20.5} & 19.2 & \underline{9.2} & \underline{12.6} & \underline{32.4} & \underline{22.6} \\
PerceptionLM~\cite{cho2025perceptionlm} & 64 & 8B & 35.7 & 14.8 & 15.1 & 0.0 & 0.0 & 0.0 & 10.9 \\
Qwen2.5-VL~\cite{bai2025qwen2} & 64 & 7B & 40.8 & 16.7 & 17.2 & 0.0 & 0.0 & 0.0 & 12.5 \\
Qwen3-VL~\cite{bai2025qwen3} & 64 & 8B & \underline{42.0} & 19.3 & \underline{19.8} & 0.0 & 0.0 & 0.0 & 13.5 \\
\bottomrule
\end{tabular}}}
\vspace{-0.4cm}
\end{table}

\subsection{Experimental Results}\label{sec:exp:main:offline}
The results are reported in~\cref{tab:offlineevals}.

\noindent\textbf{Blind baseline fails.} This serves as a simple sanity check that \bname{} cannot be solved without vision. We report it using GPT-5, since it is the strongest model in our evaluation suite.

\noindent\textbf{Closed-source models dominate open-source models.} In the Interaction and Intervention benchmarks, open-source models collapse; most score zero on SPG, SI, and UI, revealing a substantial gap to real-world streaming behavior. InternVL-3.5, explicitly trained for GUI interaction and embodied agency, is the only notable exception. Closed-source models perform substantially better across these tasks, suggesting that the gap is likely data-driven. In particular, these results point to the importance of large-scale instruction-tuning data centered around interactive, agentic, and intervention-oriented streaming scenarios. Notably, Detection performance remains relatively strong across all models, indicating that current MLLMs already possess reliable event perception capabilities.

\section{Streaming Experiments}\label{ssec:streaming_exp}
Streaming evaluation is the primary focus of \bname{}. Here, we evaluate streaming VideoLLMs using our full Timeliness-F1 metric, quantifying their streaming behavior.

\noindent\textbf{Baselines.} Evaluating streaming VideoLLMs under a unified protocol remains challenging, as prior works adopt different notions of ``streaming'' and target different tasks. Several recent methods~\cite{qian2025dispider, fu2025vispeak, yao2025timechat, wang2025streambridge, yang2025streamagent} describe proactive inference but do not release evaluation code for continuous streaming evaluation. Others~\cite{song2024moviechat, di2025streaming, ning2025livevlm} focus on long-video understanding, where queries are posed only after observing the full video. While these methods stream video inputs for efficiency, they do not support streaming outputs. We also experimented with VideoLLM-online~\cite{chen2024videollm} and M4~\cite{wang2025omnimmi}, but both showed poor instruction-following and near-zero performance on our tasks. Among existing implementations, we identify four promising candidates, including two recent \textit{interaction models}~\cite{tml2026}, designed for streaming interactions, which~\bname{} is designed to evaluate:
\vspace{-0.3cm}
\begin{itemize}
    \setlength{\itemsep}{0pt}
    \setlength{\parsep}{0pt}
    \setlength{\topsep}{0pt}
    \item \textbf{StreamingVLM}~\cite{xu2025streamingvlm}, originally trained for live captioning~\cite{chen_livecc_2025}, processes unbounded streams under bounded memory and compute. To support its silence mechanism, we additionally prompt it to emit a single ``no'' token when no response is required.
    \item \textbf{MMDuet2}~\cite{wang2025mmduet2} is post-trained for proactive streaming inference using a ``NO REPLY'' token. However, its unbounded context causes out-of-memory failures on long videos. We therefore evaluate two adaptations: \textbf{MMDuet2+KVflush} uses a sliding window and clears the KV cache at each query, limiting inference to a single active query. \textbf{MMDuet2+StreamingVLM} adopts StreamingVLM's sliding-window design and stores multi-queries in a textual cache.
    \item \textit{Interaction Models}~\textbf{MiniCPM-o 4.5}~\cite{cui2026minicpm} and \textbf{JoyAI-VL-Interaction}~\cite{yao2026joyai} are post-trained for proactive streaming interactions. We evaluate both using their default inference settings, excluding audio inputs for MiniCPM-o 4.5.
\end{itemize}
\vspace{-0.2cm}
Both StreamingVLM and MMDuet2 use Qwen2.5-VL~\cite{bai2025qwen2} as their base MLLM. For comparison, we also evaluate base \textbf{Qwen2.5-VL} by independently prompting each frame without KV retention. Finally, we evaluate \textbf{Ctrl-SPOT}, our training-free streaming controller~(\cref{sec:asynkv}), using base Qwen2.5-VL~\cite{bai2025qwen2} and InternVL-3.5~\cite{wang2025internvl3}, the strongest open-source perception model in~\cref{tab:offlineevals}.

\noindent\textbf{Evaluation Metrics.} We evaluate all streaming models using both $\mathcal{T}$-score and T-F1 at $K=5$ (T-F1@5), where $K$ denotes the occupancy budget per slot. See~\cref{supp:sec:K} for a discussion on $K$. All models are evaluated at an output frequency of 1 Hz~\ie~one prediction per second regardless of the input FPS. Since full-video streaming may contain multiple (concurrent) active queries, each query is assigned a unique tag [$Q_n$]. Responses are instructed to reuse the same tag, enabling response-query matching, with a fallback to the temporally closest query when tags are missing.

\subsection{\mname{}: A streaming controller for MLLMs}\label{sec:asynkv}
Offline MLLMs achieve strong instruction following and broad generalization when given complete visual context~\cite{bai2025qwen2, bai2025qwen3, wang2025internvl3, zhang2025videollama}. We find that adapting offline MLLMs to streaming without retraining introduces two challenges. First, most MLLMs use a front-loaded prompt structure where all frames precede the query; in a proactive setting, the query arrives first, introducing a train-test mismatch that degrades instruction following as context grows. Second, once an event occurs, the model continues responding at subsequent frames as evidence remains in context, causing \textit{response echoing}. Both failures stem from the same issue: queries are stateless across frames, with no mechanism to track prior responses. To address this, we introduce \textbf{\mname{}}, a streaming controller with:
\begin{equation}\small
\underbrace{\text{[sys]}}_{\text{sink}}
\;
\underbrace{[t_1\text{:}\,\widetilde{s}_1]\cdots[t_k\text{:}\,\widetilde{s}_k]}_{\text{episodic memory}}
\;
\underbrace{[\mathbf{F}^{Q_1}_{1:K}]\,[\mathbf{F}^{Q_2}_{1:K}]}_{\text{retrieved KV}}
\;
\underbrace{[t',Q_n\text{:}\,r_{t'}]}_{\text{answer memory}}
\;
\underbrace{[\mathbf{F}_{t-N_S:t}]}_{\text{short-term KV}}
\;
\underbrace{[t,Q_1]\,[t,Q_2]}_{\text{active queries}} \Rightarrow \text{LLM}
\label{eq:AsynKV}
\end{equation}
\noindent\textbf{i) Persistent Query State.}
Each active query is assigned a unique tag~\texttt{[$Q_n$]} with timestamp~\texttt{[$t=\tau$]} and appended to the visual context (see~\cref{eq:AsynKV}). When the model produces a non-silent response, it is similarly tagged and timestamped, then prepended to the visual KV prefix as \texttt{[$t=t_1$][$Q_n$]$r_{t_1}$}. The model therefore observes its own latest response in context on every subsequent frame, suppressing redundant outputs for the same query in close temporal proximity. Queries expire two minutes after being posed. Multiple queries may be active simultaneously and are handled in FIFO order.

\noindent\textbf{ii) Long-short term KV~($\mathbf{F}$).}
\mname{} maintains two KV memories at different temporal scales.
Short-term memory stores a sliding window of the most recent $N_S$ frames while long-term memory stores up to past $N_L$ frames. Following~\cite{di2025streaming}, long-term KV is offloaded to CPU while short-term KV remains on GPU for fast decoding. When a query is posed, top-$K$ frame KVs are retrieved from CPU using cosine similarity and inserted before the short-term KV~(\cref{eq:AsynKV}), preserving arrow of time. Following~\cite{xu2025streamingvlm, pang2026decouple}, KVs are stored pre-RoPE with 3D contiguous MRoPE applied during decoding.

\noindent\textbf{iii) Scaling dead-time compute.}
Once response echoing is suppressed, FPs decrease and the model remains silent when no queries are active. We refer to these intervals as~\underline{dead-time}, a structural property of continuous streaming inference previously hidden by pause-ask-play protocols. These intervals provide an opportunity for additional computation without affecting response latency. SPOT-Bench videos contain $\sim$60$\%$ dead-time on average. During dead-time, the model performs episodic summarization every $N_S$ seconds over non-overlapping windows. These summaries are inserted after the sinked system prompt with timestamps. This GPU-side decoding step takes $\sim$0.3s and does not drop incoming frames at 2~fps. A lightweight CPU job manages the bounded long-term memory.  A pair of consecutive frames with the highest cosine similarity is identified and the older one is discarded. Using cosine similarity for both retrieval and compression improves diversity of retrieved frames and reduces redundant context compared to ReKV.

\begin{table}[t]
\centering
\setlength{\tabcolsep}{2mm}{
\caption{\textbf{Evaluation of streaming models on \bname{}} at 1 Hz. Each cell reports Timeliness-score@5 / Timeliness-F1@5. \textbf{Post-Train?} indicates whether the model was post-trained with silence tokens for proactive VideoQA. $^\dagger$uses Qwen3-VL-4B for long-term memory summarization.}
\label{tab:streamevals}
\resizebox{\linewidth}{!}{
\begin{tabular}{@{}l c c c cc cc cc c@{}}
\toprule
\textbf{Model} & \textbf{\#} & \textbf{Post} & \textbf{Model} &
\multicolumn{2}{c}{\textbf{Detection}} &
\multicolumn{2}{c}{\textbf{Interaction}} &
\multicolumn{2}{c}{\textbf{Intervention}} &
\textbf{Macro} \\
\cmidrule(lr){5-6}\cmidrule(lr){7-8}\cmidrule(lr){9-10}
& \textbf{Frames} & \textbf{Train?} & \textbf{Size} & \textbf{ABD} & \textbf{PNR} & \textbf{SQA} & \textbf{SPG} & \textbf{SI} & \textbf{UI} & \textbf{Avg.} \\
\midrule

\rowcolor{headergray}
\multicolumn{11}{l}{\textbf{Offline MLLM}} \\
\multirow{4}{*}{\; Qwen2.5-VL~\cite{bai2025qwen2}}
& 1 & \xmark & 7B & 34.6/11.8 & 26.2/9.0 & 3.6/1.5 & 2.2/0.8 & 0.0/0.0 & 0.0/0.0 & 11.1/3.9 \\
& 8 & \xmark & 7B & 41.8/10.0 & 31.7/8.1 & 6.0/3.6 & 3.2/1.2 & 0.0/0.0 & 0.0/0.0 & 13.8/3.8 \\
& 16 & \xmark & 7B & 42.5/9.2 & 33.6/7.8 & 7.5/2.9 & 4.0/1.2 & 0.0/0.0 & 0.0/0.0 & 14.6/3.5 \\
\midrule

\rowcolor{headergreen}\multicolumn{11}{l}{\textbf{Streaming VideoLLMs}} \\
\; StreamingVLM~\cite{xu2025streamingvlm} & 2fps & \cmark & 7B & 4.8/3.9 & 4.4/2.5 & 31.6/0.9 & 1.2/0.3 & 3.2/0.7 & 2.4/0.5 & 7.9/1.5 \\
\; MMDuet2~\cite{wang2025mmduet2} & & & & & & & & & & \\
\qquad + KVflush & 2fps & \cmark & 3B & 13.9/14.2 & 4.1/5.1 & 37.8/5.3 & 2.6/3.0 & 0.0/0.0 & 0.0/0.0 & 9.7/4.6 \\
\qquad + StreamingVLM & 2fps & \cmark & 3B & 11.5/13.8 & 0.9/1.6 & 31.0/4.0 & 1.0/2.2 & 1.2/2.3 & 16.2/6.2 & 10.3/5.0 \\
\; MiniCPM-o 4.5~\cite{cui2026minicpm} & 2fps & \cmark & 9B & 11.5/15.4 & 12.1/14.3 & 7.2/6.0 & 3.5/4.9 & 0.0/0.0 & 20.8/7.2 & 9.2/8.0 \\
\; JoyAI-VL-Interaction~\cite{yao2026joyai} & 2fps & \cmark & 8B+4B$^\dagger$ & 15.3/17.0 & 32.2/15.6 & 28.1/6.2 & 9.0/\textbf{6.6} & 4.0/2.5 & 24.2/\textbf{10.0} & 18.8/9.7 \\
\addlinespace[0.6ex]
\arrayrulecolor{gray!80}\cmidrule(lr){1-11}\arrayrulecolor{black}
\; \textbf{\mname} & & & & & & & & & & \\
\qquad w/ Qwen2.5-VL & 2fps & \xmark & 7B & 40.2/\textbf{22.6} & 36.8/\textbf{19.5} & 30.8/\textbf{18.0} & 1.2/1.0 & 0.0/0.0 & 0.0/0.0 & 18.2/10.2 \\
\qquad w/ InternVL-3.5 & 2fps & \xmark & 8B & 40.8/20.5 & 37.7/15.2 & 25.6/17.8 & 5.4/4.6 & 16.6/\textbf{5.4} & 40.6/6.8 & 27.8/\textbf{11.7} \\
\bottomrule
\end{tabular}}}
\vspace{-0.3cm}
\end{table}

\subsection{Benchmark Results \& Analysis}\label{sec:exp:main:streaming}
Refer to~\cref{tab:streamevals}. T-F1 exposes model failures previously invisible under ``pause-ask-play'' evaluation.

\noindent\textbf{Larger context results in response echoing.} Naively deploying a strong MLLM like Qwen2.5-VL frame-by-frame incurs very high latency, but that is not the only issue. Scaling context improves event perception ($\mathcal{T}$-score), but degrades T-F1 as larger temporal windows keep evidence active for longer, causing repeated responses (echoing) across nearby frames. Beyond echoing, it also produces frequent FPs during dead-time~\ie~responding even when no relevant event is occurring.

\noindent\textbf{StreamingVLM over-responds.} StreamingVLM is trained for query-agnostic live commentary. As a result, it responds very frequently, producing poor precision and the lowest T-F1 overall. 

\noindent\textbf{Silence post-training reduces FPs but increases FNs.}
Post-training for proactive interaction improves performance on \bname{}, but introduces a tradeoff. Both MMDuet2 adaptations suppress over-responses through silence post-training, but become less responsive, missing more valid response slots and resulting in low T-F1. Recent interaction models, MiniCPM-o 4.5 and JoyAI-VL-Interaction, benefit from more sophisticated, large-scale post-training, retaining stronger event perception (higher $\mathcal{T}$-score) and improving Detection T-F1. However, missed responses continue to limit their performance.

\noindent\textbf{Interaction models remain far from reliable real-world assistance.}
Large-scale post-training improves response generation, as reflected in Interaction and Intervention tasks, where MiniCPM-o 4.5 and JoyAI-VL-Interaction outperform earlier streaming baselines. However, response timing itself remains challenging, as observed from Detection, and generating appropriate responses introduces another layer of complexity. Performance deteriorates further on tasks requiring long-horizon planning (SPG, SI), highlighting the substantial gap between current interaction models and reliable proactive visual assistants.

\noindent\textbf{\mname{} retains perception while improving streaming behavior.} With Qwen2.5-VL, \mname{} improves overall T-F1 from 3.9\% to 10.2\%, while retaining strong Detection performance. Unlike silence post-training, the gains come without sacrificing $\mathcal{T}$-score, suggesting that training-free harnesses can mitigate over-response without weakening perception.

\noindent\textbf{\mname{} generalizes across backbone families.} Using InternVL-3.5 as the backbone further improves Interaction and Intervention performance, establishing a strong open-source baseline for future streaming model research on \bname{}.

\noindent\textbf{Scaling dead-time compute.}
Dead-time is an interesting aspect of \bname{}. As shown in~\cref{tab:scaling_ablation}, progressively adding memory components increases compute (GFLOPs averaged per video-second), while keeping response latency bounded ($\sim0.5$s), enabling real-time responses at 2 fps without buffering or dropping frames. For fair comparison, GFLOPs are measured on a PNR subset under a controlled setting (1 fps, 64 tokens per frame) to isolate the computational cost of each memory component.

\noindent\textbf{\mname{}'s strong transfer to Retrospective VideoQA}.
Results are provided in~\cref{tab:retroqa}. We particularly evaluate on real-time VideoQA, where queries concern recently observed events, as proactive queries can be viewed as repeatedly answering such queries over a stream. \mname{} requires no modification for this setting and can be used directly, with its stronger memory management yielding better performance.

\begin{table}[t]
\centering
\vspace{0.2cm}

\begin{minipage}[t]{0.49\linewidth}
\centering
\caption{\textbf{RetrospectiveQA} Top-1 accuracy.}
\label{tab:retroqa}
\setlength{\tabcolsep}{0.5mm}
\renewcommand{\arraystretch}{1.1}
\resizebox{\linewidth}{!}{
\begin{tabular}{lcc}
\toprule
\textbf{Model} & \shortstack{\textbf{StreamingBench}\\\textbf{Real-time~\cite{lin2024streamingbench}}} & \shortstack{\textbf{OVO-Bench}\\\textbf{Real-time~\cite{niu2025ovo}}} \\
\midrule
StreamAgent~\cite{yang2025streamagent} & 74.3 & 61.3 \\
TimeChat-online~\cite{yao2025timechat} & 75.4 & 61.9 \\
StreamingVLM~\cite{xu2025streamingvlm} & - & 62.0 \\
StreamForest~\cite{zeng2025streamforest} & 77.3 & 61.2 \\
StreamBridge~\cite{wang2025streambridge} & 77.0 & 71.3 \\
\rowcolor{headergreen}
\textbf{\mname{}} & \textbf{79.9} & \textbf{71.6} \\
\bottomrule
\end{tabular}}
\end{minipage}
\hfill
\begin{minipage}[t]{0.47\linewidth}
\centering
\caption{\textbf{Scaling ablation.} Effect of memory components on performance (T-F1), GFLOPs/sec. and latency on a single H100.}
\label{tab:scaling_ablation}
\setlength{\tabcolsep}{0.1mm}
\renewcommand{\arraystretch}{1.1}
\resizebox{\linewidth}{!}{
\begin{tabular}{lcccc}
\toprule
\textbf{Model} & \textbf{ABD} \quad & \textbf{PNR} \quad & \textbf{GFLOPs} \quad & \textbf{Latency(s)} \\
\midrule
StreamingVLM~\cite{xu2025streamingvlm} & 3.9 & 2.5 & 1301 & 0.26 \\
\rowcolor{headergray}
Qwen2.5-VL~\cite{bai2025qwen2} & 9.2 & 7.8 & 4426 & 2.56 \\
\quad + short-term KV & 19.1 & 17.6 & 1640 & 0.22 \\
\quad + long-term KV & 21.9 & 19.0 & 2708 & 0.46 \\
\quad + episodic sum. & 22.6 & 19.5 & 3660 & 0.51 \\
\bottomrule
\end{tabular}}
\end{minipage}
\vspace{-0.4cm}
\end{table}

\vspace{-0.2cm}

\section{Conclusion}
\vspace{-0.2cm}
In this work, we move beyond the pause-ask-play paradigm of Online VideoQA and introduce \bname{}, the first benchmark for proactive, multi-turn, multi-response streaming evaluation. Our results reveal a clear gap between closed- and open-source models on Interaction and Intervention tasks, highlighting the need for higher-quality streaming instruction tuning data and training methods for assistive and agentic scenarios. We hope SPOT-Bench motivates more targeted data collection efforts in this space. We further identify dead-time as a new axis for scaling streaming VideoLLMs without affecting response latency. Together, \bname{} and Timeliness-F1 provide a foundation for future research toward general-purpose, always-on, real-time assistants.

\textbf{Acknowledgements.} This research / project is supported by the National Research Foundation, Singapore, under its NRF Investigatorship (NRFI Award No.: NRF-NRFI11-2026-0006). We thank Dishantika Chatterjee, Aryan Jain, and Eyl\"ul Pelin K{\i}l{\i}\c{c} for their time and help with the benchmark annotation and review process. We also thank Tanuj Sur and Junbin Xiao for helpful discussions.

{
    \bibliography{main}
    \bibliographystyle{plainnat}
}

\clearpage

\appendix

\begin{center}
{\Large\bfseries Appendix}
\end{center}

{\noindent\large\bfseries Table of Contents}
\vspace{0.3em}
\hrule
\vspace{0.8em}

{\small
\noindent\textbf{\ref{app:limitations}\quad Limitations}
\dotfill\pageref{app:limitations}\\

\noindent\textbf{\ref{app:impacts}\quad Broader Impacts}
\dotfill\pageref{app:impacts}\\

\noindent\textbf{\ref{app:related}\quad Extended Related Work}
\dotfill\pageref{app:related}\\

\noindent\textbf{\ref{supp::sec:spot}\quad SPOT-Bench: Additional Details}
\dotfill\pageref{supp::sec:spot}\\

\hspace*{1.2em}\ref{supp::sec:spot:detection}\quad Detection as a Perception Benchmark
\dotfill\pageref{supp::sec:spot:detection}\\

\hspace*{1.2em}\ref{supp::sec:spot:interaction}\quad Interaction \& Intervention as Diagnostic Benchmarks
\dotfill\pageref{supp::sec:spot:interaction}\\

\hspace*{1.2em}\ref{supp::ssec:annotation}\quad Data Collection \& Annotation Pipeline
\dotfill\pageref{supp::ssec:annotation}\\

\hspace*{1.2em}\ref{supp::ssec:stats}\quad SPOT-Bench Statistics
\dotfill\pageref{supp::ssec:stats}\\

\noindent\textbf{\ref{app:metric}\quad Streaming Evaluation Metric}
\dotfill\pageref{app:metric}\\

\hspace*{1.2em}\ref{supp:ssec:human-study}\quad Controlled Human Study
\dotfill\pageref{supp:ssec:human-study}\\

\hspace*{1.2em}\ref{supp::ssec:t-f1}\quad Timeliness-F1
\dotfill\pageref{supp::ssec:t-f1}\\

\noindent\textbf{\ref{app:exp}\quad Additional Experiments \& Analysis}
\dotfill\pageref{app:exp}\\

\hspace*{1.2em}\ref{app:implementation}\quad Ctrl-SPOT Implementation Details
\dotfill\pageref{app:implementation}\\

\hspace*{1.2em}\ref{app:ablations}\quad Ablations \& Analysis
\dotfill\pageref{app:ablations}\\

\noindent\textbf{\ref{app:exemplars}\quad Task Exemplars}
\dotfill\pageref{app:exemplars}
}

\vspace{0.8em}
\hrule
\newpage

\section{Limitations}\label{app:limitations}
\bname{} currently focuses on visual-only streaming inputs; extending to audio-visual streams (e.g., spoken instructions or ambient sound cues) is a natural direction for future work. While our Detection benchmark is large-scale, the Interaction and Intervention splits are intentionally smaller, reflecting the long-tail distribution of such events in the real world -- scale here trades off against the precision and objectivity of annotations (very high inter-annotator agreement). Our evaluation of open-ended responses relies on LLM-as-Judge (GPT-5-mini), which, while standard practice, introduces a dependency on closed-source APIs; the Detection tasks deliberately avoid this by using single-token outputs.

\section{Broader Impacts}\label{app:impacts}
Always-on streaming assistants can assist blind individuals navigating unsafe environments, coach athletes and surgeons by flagging errors in real time, and guide users learning new procedural skills like cooking or assembly. SPOT-Bench is a step toward making such systems reliable. As with any assistive technology, false positives and missed interventions carry real consequences; deployment in safety-critical settings should involve human oversight.

\section{Extended Related Work}\label{app:related}
\noindent\textbf{Video Large Language Models} (VideoLLMs) extend large language models \cite{achiam2023gpt, dubey2024llama, yang2025qwen3} beyond text by pairing them with strong visual encoders~\cite{radford2021learning, zhai2023sigmoid} and training on large-scale image-video-text data. VideoLLMs~\cite{maaz2023video, zhang2024video, zhang2023video, cho2025perceptionlm} have greatly advanced open-ended video understanding.  However, they are primarily designed for offline settings, where the full video can be processed and queried to answer user prompts. Accordingly, research has focused on enhancing instruction following through scaling pretraining~\cite{zhang2024video, li2024llava, bai2025qwen2}, proposing new architectures~\cite{liu2024improved, wang2024qwen2} or improving task generalization~\cite{li2024llavainter, li2024llava} and reasoning~\cite{feng2025video, sun2025video}. This has resulted in a suite of robust offline models. However, they remain ill-suited for streaming video understanding, where inference must be made continuously under partial observations and queries may be asynchronous or entirely absent, requiring models to handle temporal uncertainty and real-time decision making.

\section{SPOT-Bench: Additional Details}\label{supp::sec:spot}
\subsection{Detection as a Perception Benchmark}\label{supp::sec:spot:detection}
Detection is the core benchmark of SPOT-Bench. It isolates the perception and anticipative abilities a streaming VideoLLM must have before it can behave proactively in any meaningful way. ABD and PNR minimize the burden of language generation and test only one thing: whether the model responds at the correct moment as events unfold. In this sense, detection evaluates instruction following from a perception standpoint~\ie~responding timely to an instruction. Predictions are single-token outputs such as ``start'', ``end'', or ``now''. This enables direct semantic matching without requiring LLM-as-Judge evaluation. As a result, large-scale evaluation does not depend on expensive frontier APIs, making benchmarking inclusive and practical to run at scale across our 662 detection videos. The single-token design also avoids response completion latency in autoregressive models and provides a fair basis for comparing autoregressive and non-autoregressive systems.

Detection is constructed to mirror the temporal complexity of real-world streaming scenarios. The videos are long (Avg: 3 mins 40 secs; Max: 53 mins), contain multiple turns, often require multiple responses per turn, include concurrent queries, and span both egocentric and exocentric viewpoints. These settings present a variety of perception challenges that proactive assistants must handle. Unlike the Interaction and Intervention benchmarks, Detection yields reasonable scores across a wide range of models (\cref{tab:offlineevals}), showing that it is difficult but still approachable with present SOTAs. Human annotators also show high agreement (\cref{fig:supp::intervals}), confirming that the gold annotations in Detection are objective and well-defined.

\subsection{Interaction \& Intervention as Diagnostic Benchmarks}\label{supp::sec:spot:interaction}
Interaction and Intervention are novel benchmark constructions introduced in SPOT-Bench. Here, the goal is not to maximize quantity, but to reflect real-world distributions, where such events lie in the long tail. Accordingly, we design these as hard challenge splits to be future-proof and expose realistic failure modes. Intervention tasks, in particular, can exhibit subjectivity in their response intervals. Our curation therefore focuses on a high-precision subset, resulting in naturally objective response intervals as reflected in~\cref{tab:supp:anno-agreement}, which shows strong annotator agreement (low MAD/IQR). Results in~\cref{tab:offlineevals} further show that strong event perception alone does not translate into competent assistive behavior. These benchmarks therefore probe a distinct capability: whether a model can provide meaningful assistance in real-time.

\subsection{Data Collection \& Annotation Pipeline}\label{supp::ssec:annotation}
Our benchmark requires gold annotations marking the exact timestamp(s) when a query should be answered. At first glance, one might consider repurposing existing Online VideoQA datasets. These datasets feature questions, answers, and their corresponding asking timestamps, which are equal to their response time~\cite{wang2025streambridge}. In principle, these could be adapted by shifting the asking time backward so that the response follows it. However, this assumption quickly breaks down, as many questions are either \textit{retrospective} (\eg~``Why is Mr.~Bean shocked now?''~\cite{lin2024streamingbench}; ``Which object did the person eat before they threw the clothes?''~\cite{niu2025ovo}) or \textit{static} (\eg, ``What is the person holding right now?''~\cite{lin2024streamingbench}; ``What is displayed on the house number plate?''~\cite{niu2025ovo}), offering no signal of \textit{when} the referenced event actually occurred. Consequently, identifying the correct response timestamp demands explicit manual annotation.

\noindent\textbf{Data Collection.} We curate datasets with human-annotated action segments and rich metadata, which provide strong priors to make our annotation process more efficient. We source from both egocentric and exocentric video datasets, as streaming applications differ significantly across these viewpoints and should be equally explored. 
\cref{tab:task_stats} lists the source datasets used for each benchmark category. We use only the validation splits of these datasets to maintain consistency with their original evaluation protocols and to minimize the risk of data contamination. For PNR, we take the FHO split from Ego4D and the REC split from OVO-Bench~\cite{niu2025ovo}, which is sourced from THUMOS’15. For SQA, we use the CRR split from OVO-Bench, which is sourced from MovieNet.

\begin{figure}[t]
\centering
\vspace{0.3cm}
\includegraphics[width=\linewidth]{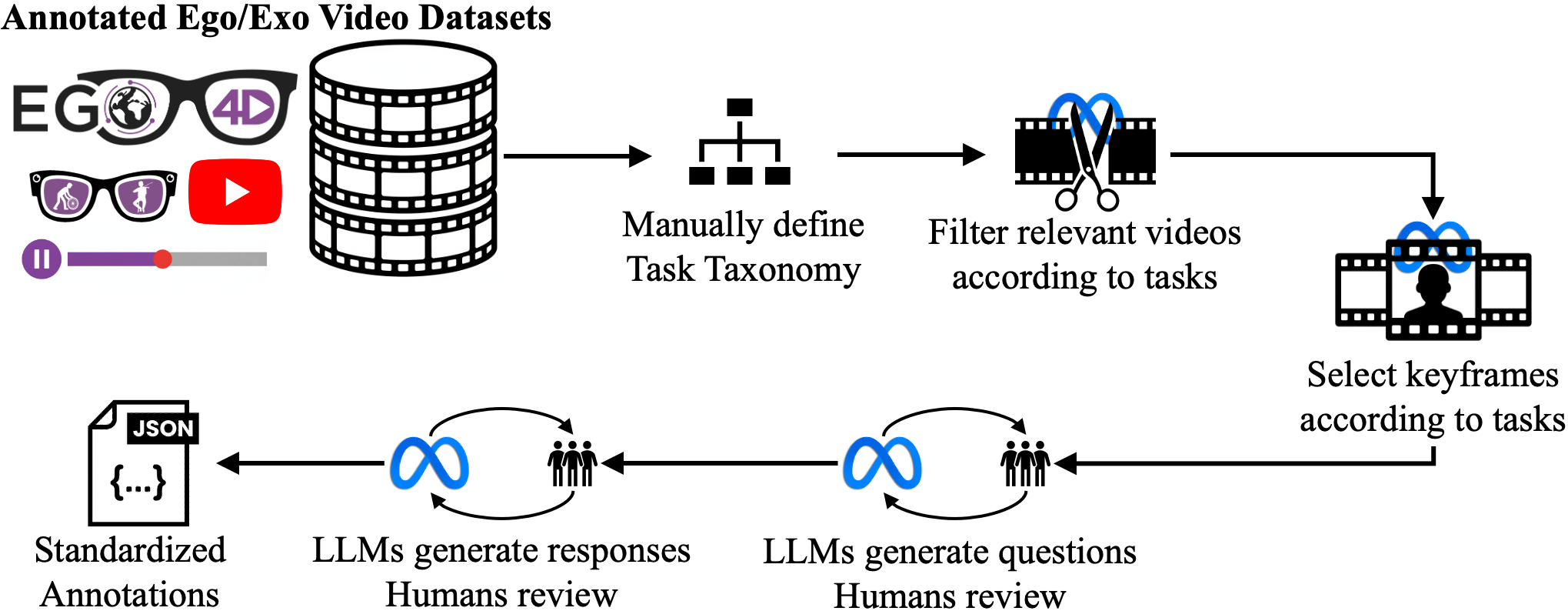}
\caption{\textbf{Our annotation pipeline.} We first define a task taxonomy and probe curated datasets for relevant videos. After filtering according to the taxonomy, we generate ground-truth annotations in a semi-automatic process where LLMs assist human annotators.}
\vspace{-5mm}
\label{fig:supp::annotation}
\end{figure}

\noindent\textbf{Data Filtering.}
After defining our benchmark tasks and taxonomy, we filter the collected videos to retain only segments relevant to each task. To assess task relevance, we use Llama-3.3-70B-Instruct~\cite{dubey2024llama} to screen segment metadata and ground-truth annotations of the corresponding source datasets: EgoExo4D and HT-Step for ABD, Ego4D for PNR, and EgoBlind for UI. The remaining datasets either have incomplete metadata or are small in scale, which makes manual screening more efficient. 
We then manually remove redundant clips and videos with highly repetitive actions, retaining a compact yet diverse set of videos. Finally, Llama is used once more to generate candidate~\textit{keyframes} - critical temporal points such as action boundaries or PNR frames (see \cref{fig:supp::annotation}). Task specific prompts are provided in Prompts~\ref{prompt:ABD},\ref{prompt:PNR}.

\noindent\textbf{Annotation Generation.}
The filtered videos feature multiple relevant segments, spanning a diverse range of actions. Next is to convert every segment into a meaningful question-response turn. For EgoExo4D, HT-Step, Ego4D, and HoloAssist, since these datasets already provide some ground-truth event segments and video-level metadata, we use Llama-3.3-70B-Instruct~\cite{dubey2024llama} to generate initial queries based on such annotations. Once all per-segment queries are generated for a video, the same model is prompted to identify possible referential links across them, allowing temporal and contextual continuity between adjacent turns. A human annotator then reviews all generated queries and referential links, and assigns timestamps for the keyframes, guided by the original ground-truth bounds and cleaned metadata. This semi-automated process significantly improves annotation efficiency while maintaining temporal precision. A second annotator subsequently reviews the full question-response annotations for accuracy and consistency. We observe high inter-annotator agreement in detection tasks, with comparatively lower agreement in interaction and intervention tasks. The latter is expected, given the inherent ambiguity of point-wise annotations, although most responses still fall within a consistent temporal interval. This annotation ambiguity is addressed through our proposed streaming evaluation metric~(\cref{sec:ssec::evaluation}).

\subsection{\bname{} Statistics}\label{supp::ssec:stats}
\begin{figure}[t]
\centering
\includegraphics[width=\linewidth]{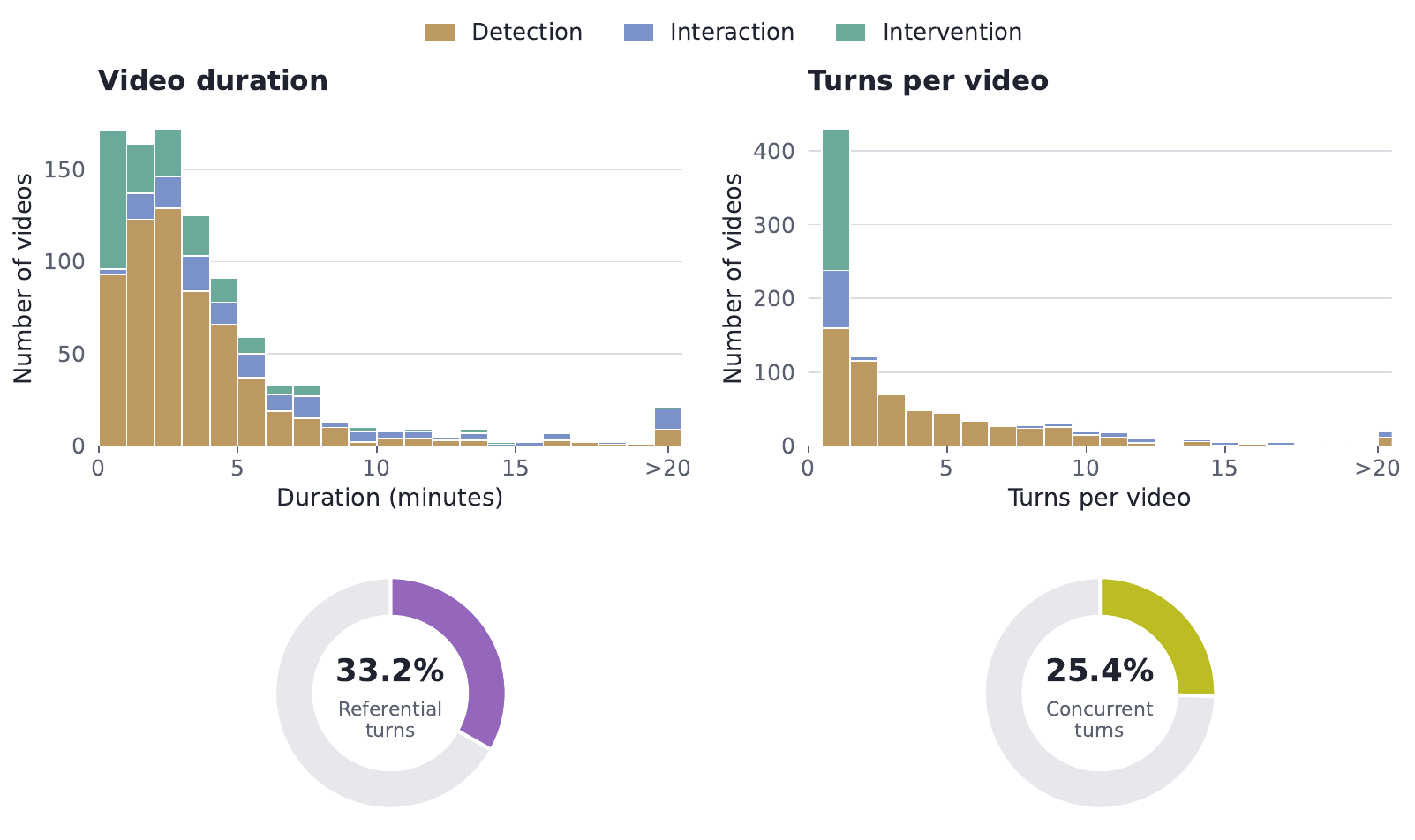}
\caption{\textbf{Additional \bname{} Statistics.} Overview of video durations, turn counts and our special turn types: Referential and Concurrent.}
\label{fig:supp::stats}
\end{figure}

SPOT-Bench comprises 940 videos totaling 61 hours of streaming footage, with 3837 turns and 5093 responses.
Each video averages 4~min~1~secs, with 4.08 turns per video and 1.33 responses per turn. To our knowledge, this is the first benchmark to enable multi-turn, multi-response evaluation of streaming models. Task statistics are reported in~\cref{tab:task_stats}. We visualize additional statistics in~\cref{fig:supp::stats}.

Additionally, SPOT-Bench introduces several novel evaluation settings. Our Intervention tasks, especially the unsolicited variant, target safety-critical proactive behavior that remains unexplored in prior online or streaming datasets. We also provide referential and concurrent QA scenarios: \underline{referential} queries' responses depend on previous turns, while \underline{concurrent} queries allow multiple active queries, reflecting overlapping conversational events. Beyond scale, SPOT-Bench offers a diverse suite of streaming challenges that require not only timely responses, but also persistent memory, multi-turn instruction following, continuous progress monitoring, and proactive intervention.

\section{Streaming Evaluation Metric}\label{app:metric}

\begin{figure}[t]
\centering
\begin{minipage}[t]{0.65\textwidth}
    \vspace{0pt}
    \centering
    \includegraphics[width=\linewidth]{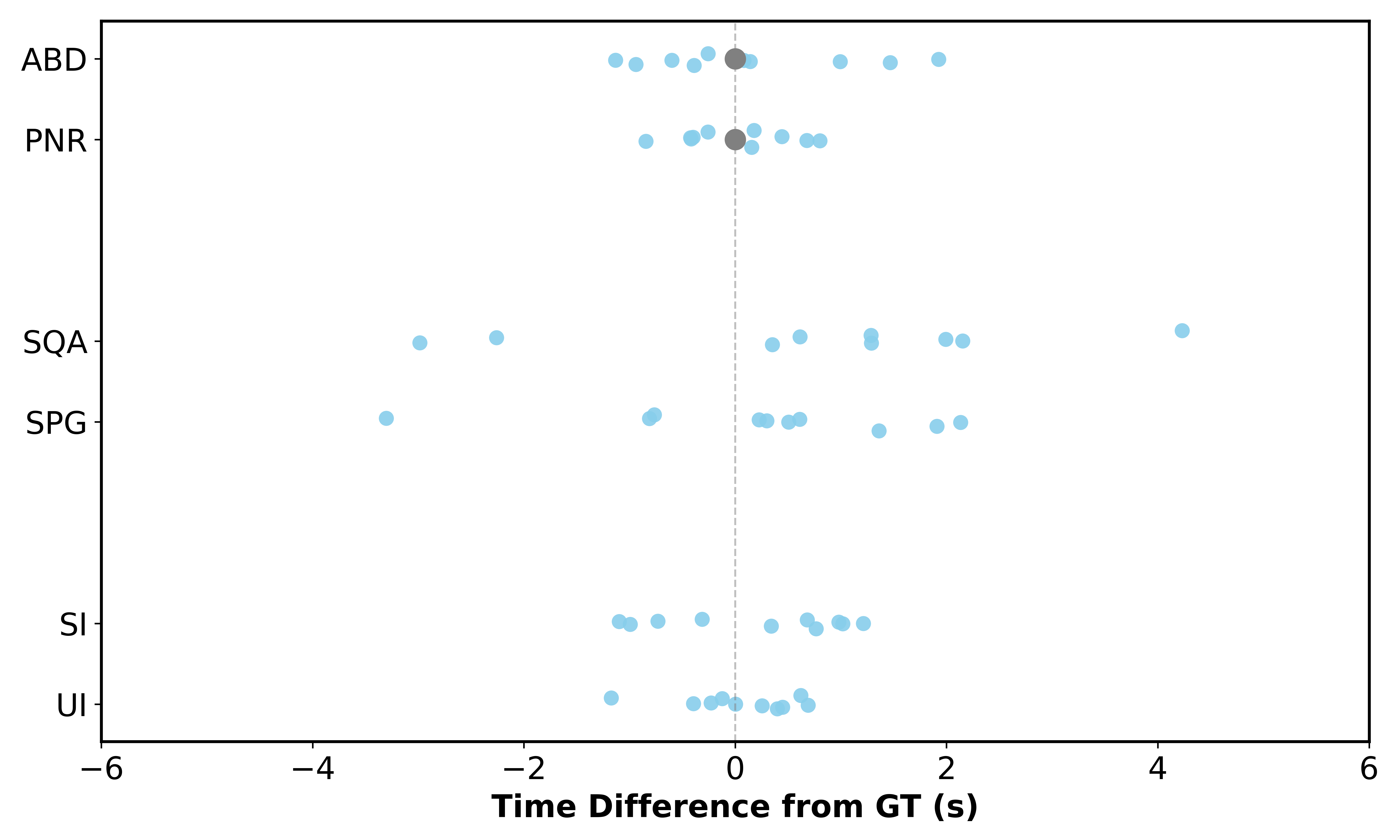}
    \caption{\textbf{Human Preference Intervals.} Gold intervals derived from our controlled human study with 10 annotators. Blue dots show annotator response times relative to $0$s~(\cref{sec:ssec::evaluation}). For detection tasks (PNR and ABD), gray dots mark the precise keyframe timestamps provided by the source datasets.}
    \label{fig:supp::intervals}
\end{minipage}
\hfill
\begin{minipage}[t]{0.32\textwidth}
    \vspace{0pt}
    \centering
    \setlength{\tabcolsep}{2.5mm}{
    \resizebox{\linewidth}{!}{
    \begin{tabular}{@{}lcc@{}}
    \toprule
    \textbf{Task} & \textbf{MAD (s)} & \textbf{IQR (s)} \\
    \midrule
    \rowcolor{headergray}
    \multicolumn{3}{l}{\textbf{Detection}} \\
    \qquad ABD & 0.4 & 1.0 \\
    \qquad PNR & 0.2 & 0.5 \\
    \addlinespace[0.6ex]
    \rowcolor{headergray}
    \multicolumn{3}{l}{\textbf{Interaction}} \\
    \qquad SQA & 1.1 & 2.5 \\
    \qquad SPG & 0.6 & 1.5 \\
    \addlinespace[0.6ex]
    \rowcolor{headergray}
    \multicolumn{3}{l}{\textbf{Intervention}} \\
    \qquad SI & 0.5 & 1.0 \\
    \qquad UI & 0.3 & 0.5 \\
    \bottomrule
    \end{tabular}}}
    \captionof{table}{\textbf{Inter-annotator agreement} for human preference intervals is reported as mean absolute deviation (MAD) in secs. Interquartile Range (IQR) in secs directly corresponds to the timeliness thresholds chosen for each task.}
    \label{tab:supp:anno-agreement}
\end{minipage}
\end{figure}

\subsection{Controlled Human Study.}\label{supp:ssec:human-study}
\textbf{Setup.} We recruited over 20 annotators. Annotators were shown the live video and provided with the queries and asked: ``At what point in the video would you expect an assistant to respond to this question? Mark the earliest moment at which you have sufficient information to give a correct answer.'' Task-specific guidance was provided for each category. From the 20, 10 met our quality bar. These annotators marked acceptable response intervals across representative samples for each task. We consider 10 annotators consistent with existing Online VideoQA benchmarks (StreamingBench, ProactiveVideoQA and OmniMMI use 5, 4, and 2 annotators respectively). After filtering outliers, we retain the eight annotations with the lowest intra-task variance (80\% of qualified annotators) to establish human preference intervals and compute the average response timestamps for each task. For Detection tasks (PNR, ABD), we sort videos in ascending order by number of turns and responses, and sample the 100 most densely annotated videos per task (\ie~200 total). For Interaction and Intervention tasks, we use all available videos across tasks.

\noindent\textbf{Outcomes.}~\cref{fig:supp::intervals} visualizes the preference intervals obtained from our controlled human study. Agreement statistics are reported in~\cref{tab:supp:anno-agreement} using MAD (Mean Absolute Deviation, in seconds) and IQR (Interquartile Range, in seconds). We observe that \textit{agreement correlates with task granularity}. Detection tasks (ABD, PNR) show the highest agreement due to their precise temporal definitions. Interaction tasks (SQA, SPG) show lower agreement, reflecting the open-ended nature of assistive queries. Notably, the Intervention benchmarks (SI, UI) achieve high agreement despite inherent subjectivity, highlighting the effectiveness of our curation pipeline in constructing a high-precision subset with more objective events. 

\begin{table}[h]
\vspace{-0.45cm}
\centering
\setlength{\tabcolsep}{6mm}{
\caption{\textbf{Task-specific timeliness kernel parameters.}}
\label{tab:supp:kernel_params}
\begin{tabular}{@{}lcccccc@{}}
\toprule
\textbf{Task} & \textbf{PNR} & \textbf{ABD} & \textbf{SQA} & \textbf{SPG} & \textbf{SI} & \textbf{UI} \\
\midrule
\textbf{IQR (s)} & 0.5 & 1.0 & 2.5 & 1.5 & 1.0 & 0.5 \\
$\sigma_{\mathrm{early}}$ & 0.74 & 1.48 & 3.71 & 2.22 & 1.48 & 0.74 \\
$\sigma_{\mathrm{late}}$ & 0.37 & 0.74 & 1.85 & 1.11 & 0.74 & 0.37 \\
\bottomrule
\end{tabular}}
\end{table}

\noindent\textbf{Timeliness kernel calibration.}
We derive the task-specific timeliness kernel~(\cref{fig:evaluation} right) parameters from the controlled human study rather than choosing them arbitrarily. For each task, we fit a Gaussian to the annotator response times and use the interquartile range (IQR) to define the gold interval ($t_e-t_s=2\cdot\mathrm{IQR}$). Then using the standard Gaussian-consistency relation $\sigma=\mathrm{IQR}/1.349$, we set $\sigma_{\mathrm{early}}=2\sigma$ and $\sigma_{\mathrm{late}}=\sigma$, penalizing late responses more strongly than early ones. The interval multiplier of 2 yields approximately 80\% coverage of the fitted Gaussian, consistent with retaining 80\% of annotators. Empirically, varying the interval width and early/late asymmetry does not change the relative model rankings reported in~\cref{tab:offlineevals} or~\cref{tab:streamevals}.

\subsection{Timeliness-F1}\label{supp::ssec:t-f1}

\noindent\textbf{Computing semantic matches.}
For the Detection benchmark, outputs are restricted to single tokens. Semantic matches are therefore computed using either exact match or edit distance with a 0.8 threshold, both yielding identical results. For the Interaction and Intervention benchmarks, we use GPT-5-mini as an LLM-as-Judge to determine the semantic correctness of model responses. The evaluation prompts used for these four tasks are provided in Prompts~\ref{prompt:SQA}, \ref{prompt:SPG}, \ref{prompt:SI}, and \ref{prompt:UI}.

\noindent\textbf{LLM-as-Judge validation.}
The semantic judge determines whether a temporally valid response is counted as a true positive or rejected as a false positive. To validate its reliability, we randomly sample 50 response slots from each of SQA, SPG (100 in total for Interaction) and SI, UI (100 in total for Intervention). We evaluate the same responses using GPT-5-mini, Gemini-3.5-Flash-Lite (with identical prompts as stated above), and three human judges following the same 0-5 scoring rubric. We report binary acceptance agreement and quadratic-weighted Cohen's $\kappa$ between GPT-5-mini and the median human score, alongside inter-human agreement in~\cref{tab:supp:judge-validation}. We use responses generated by GPT-5 and Gemini-3-Flash, the strongest-performing models on the benchmark, as they produce accurate and semantically diverse responses. 

\begin{table}[t]
\centering
\setlength{\tabcolsep}{2.5mm}{
\caption{\textbf{LLM-as-Judge validation against human annotations.}
$\mathcal{T}$-score@1 is reported using GPT-5-mini and Gemini-3.5-Flash-Lite as semantic judges. Agreement is measured against the median score of three human annotators, alongside human inter-rater agreement. $\kappa$ denotes quadratic-weighted Cohen's kappa.}
\label{tab:supp:judge-validation}
\resizebox{\linewidth}{!}{
\begin{tabular}{@{}ll cc cc cc@{}}
\toprule
\multirow{2}{*}{\textbf{Evaluated Model}} &
\multirow{2}{*}{\textbf{Category}} &
\multicolumn{2}{c}{\textbf{$\mathcal{T}$-score@1}} &
\multicolumn{2}{c}{\textbf{LLM--Human}} &
\multicolumn{2}{c}{\textbf{Human Inter-rater}} \\
\cmidrule(lr){3-4}
\cmidrule(lr){5-6}
\cmidrule(lr){7-8}
& &
\textbf{Gemini Judge} & \textbf{GPT Judge} &
\textbf{Agree. (\%)} & \textbf{$\kappa$} &
\textbf{Agree. (\%)} & \textbf{$\kappa$} \\
\midrule
\multirow{2}{*}{GPT-5}
& Interaction   & 45.4 & 45.9 & 97.4 & 0.96 & 98.9 & 0.96 \\
& Intervention  & 45.7 & 44.4 & 96.7 & 0.95 & 97.1 & 0.92 \\
\midrule
\multirow{2}{*}{Gemini-3-Flash}
& Interaction   & 28.0 & 28.0 & 98.3 & 0.96 & 99.1 & 0.96 \\
& Intervention  & 28.6 & 26.6 & 94.7 & 0.91 & 97.2 & 0.92 \\
\bottomrule
\end{tabular}}}
\end{table}

GPT-5-mini achieves 94.7-98.3\% binary agreement with human consensus and $\kappa=0.91$-$0.96$, closely matching inter-human agreement of 97.1--99.1\% and $\kappa=0.92$--$0.96$. Furthermore, Gemini-3.5-Flash-Lite produces nearly identical $\mathcal{T}$-score@1 across both benchmark categories and model families, indicating that the reported results are robust across different LLM judges.

\noindent\textbf{Greedy Matching for Timeliness-F1.}
\cref{alg:supp::greedy_matching} illustrates the matching process. We track four quantities during evaluation: \textit{x}TP, the sum of weighted Timeliness-scores; TP, the count of matched predictions; FP, the count of unmatched predictions; and FN, the count of unmatched slots. Precision ($\mathbf{P}$), Recall ($\mathbf{R}$), and the final Timeliness-F1 (T-F1) are computed as:
\begin{equation}\small
\mathbf{P} = \frac{\text{\textit{x}TP}}{\text{TP} + \text{FP}}, \quad
\mathbf{R} = \frac{\text{\textit{x}TP}}{\text{TP} + \text{FN}}, \quad
\text{T-F1} = 2 \cdot \frac{\mathbf{P} \cdot \mathbf{R}}{\mathbf{P} + \mathbf{R}}.
\end{equation}

\begin{algorithm}[h!]
\caption{Greedy Matching for Timeliness-F1@$K$}
\label{alg:supp::greedy_matching}
\begin{algorithmic}[1]\small
\Require 
\begin{tabular}[t]{@{}l@{}}
    Responses $\mathcal{R} = \{r_{\tau_i}\}_{i=1}^{N}$, \\
    Slots $\mathcal{G} = \{(t_{s,j}, t_{e,j}, \hat{r}_j)\}_{j=1}^{M}$, \\
    Functions $\mathcal{T}, \mathcal{S}$ \\
    Occupancy budget $K \geq 1$
\end{tabular}
\State \textbf{Context:} Single video stream $V$.
\State Sort $\mathcal{R}$ by timestamp $\tau$ ascending
\State Sort $\mathcal{G}$ by start time $t_s$ ascending
\State $\text{FP} \leftarrow 0$
\State $\mathcal{F} \leftarrow \emptyset$ \Comment{Set of closed slot indices (collected $K$ predictions)}
\State $C[j] \leftarrow 0, \quad \mathcal{T}^*[j] \leftarrow 0 \;\;\forall\, j \in \{1,\dots,M\}$ \Comment{Per-slot match count \& best $\mathcal{T}$-score}

\For{$i \leftarrow 1$ \textbf{to} $N$} \Comment{Iterate responses chronologically}
    \State $\text{matched} \leftarrow \text{False}$

    \For{$j \leftarrow 1$ \textbf{to} $M$} \Comment{Check every non-closed slot}
        \If{$j \in \mathcal{F}$} \textbf{continue} \EndIf \Comment{Skip closed slots}

        \State $\mathcal{T}_j \leftarrow \mathcal{T}(\tau_i;\, t_{s,j},\, t_{e,j})$ \Comment{Timeliness-score}
        \If{$\mathcal{T}_j = 0$} \textbf{continue} \EndIf \Comment{Not temporally valid}

        \If{$\mathcal{S}(r_{\tau_i},\, \hat{r}_j) = \text{True}$} \Comment{Semantic match (exact / GPT-as-Judge)}
            \State $C[j] \leftarrow C[j] + 1$ \Comment{Collect prediction in slot $j$}
            \State $\mathcal{T}^*[j] \leftarrow \max\!\bigl(\mathcal{T}^*[j],\; \mathcal{T}_j\bigr)$ \Comment{Update best $\mathcal{T}$-score}
            \If{$C[j] = K$}
                \State $\mathcal{F} \leftarrow \mathcal{F} \cup \{j\}$ \Comment{Close slot: budget reached}
            \EndIf
            \State $\text{matched} \leftarrow \text{True}$
        \EndIf
    \EndFor

    \If{\textbf{not} matched}
        \State $\text{FP} \leftarrow \text{FP} + 1$ \Comment{Prediction matched no slot}
    \EndIf
\EndFor

\State $\text{TP} \leftarrow \bigl|\{j : C[j] > 0\}\bigr|$ \Comment{Slots with $\geq 1$ matched prediction}
\State $\text{\textit{x}TP} \leftarrow {\displaystyle\sum_{j:\, C[j]>0}} \mathcal{T}^*[j]$ \Comment{Sum of best $\mathcal{T}$-scores}
\State $\text{FN} \leftarrow \bigl|\{j : C[j] = 0\}\bigr|$ \Comment{Slots with no matched prediction}
\State \Return $\text{\textit{x}TP},\; \text{TP},\; \text{FP},\; \text{FN}$
\end{algorithmic}
\end{algorithm}

\noindent\textbf{Influence of $K$.}\label{supp:sec:K}
\cref{alg:supp::greedy_matching} generalizes Timeliness-F1 through an occupancy budget $K$ that controls how many temporally valid, semantically correct predictions each slot may collect before closing. At $K{=}1$, each slot accepts only the first matched prediction, and its $\mathcal{T}$-score becomes the fractional TP credit; every subsequent prediction that cannot be matched to any open slot is counted as a false positive. This corresponds to the strictest user model: the assistant's first correct response is taken at face value, and all later outputs, regardless of quality, are treated as over-response. While faithful to the design philosophy that \textit{every prediction matters}, this setting can be overly punitive for current models, which often produce a correct response near the target event but not on their very first attempt. Moreover, our timeliness intervals are calibrated at the task level, whereas individual queries may have different temporal tolerances. Allowing multiple valid predictions per slot provides some robustness to this intra-task variation without changing the underlying timeliness kernel. To this end, increasing $K$ relaxes the metric along two complementary axes. First, because each slot retains the maximum $\mathcal{T}$-score among its $K$ collected predictions, a model that initially responds slightly early or late can improve its fractional credit if a better-timed correct prediction arrives within the budget. Second, the $K{-}1$ additional collected predictions are absorbed by the slot rather than counted as false positives, which improves precision. Crucially, a prediction that is not temporally or semantically valid for any slot remains a false positive regardless of $K$, preserving the interpretability of precision as a measure of over-response. From a user's perspective, $K>1$ models a more patient observer who, rather than committing to the assistant's first response, considers a short window of up to $K$ attempts and judges the assistant by its best-timed answer within that window. This is directly analogous to top-$K$ accuracy in multi-class classification, where a prediction is deemed correct if the true label appears anywhere among the model's $K$ highest-confidence outputs; here, a slot is deemed answered if at least one of the first $K$ matched predictions is well-timed, and the slot's credit reflects the best temporal alignment achieved. In practice, we set $K=5$ which balances tolerance for minor response jittering and query variation, while still penalizing persistent over-responses.

\section{Additional Experiments \& Analysis}\label{app:exp}

\subsection{\mname{} Implementation Details}\label{app:implementation}
\mname{} is implemented on top of Qwen2.5-VL-7B~\cite{bai2025qwen2} and InternVL-3.5~\cite{wang2025internvl3}. \mname{} maintains a short-term KV of $N_S{=}16$ frames on GPU. Long-term KV stores up to $N_L{=}256$ frames on CPU. For each active query, we retrieve the top-$K{=}2$ relevant frames from long-term memory. We maintain up to 5 episodic summaries, each computed over 16-second non-overlapping windows during dead-time. A query remains active for 120 seconds (for ABD, PNR and SQA that has explicit queries) and is assigned a persistent tag \texttt{[Qn]}. This tag is also used to match responses to queries when multiple (concurrent) queries are active. Answer memory stores the latest tagged response for each active query. Because the model is not trained to remain silent, we instruct it to output ``no'' as a silence token. Memory compression removes highly similar adjacent frames from external memory using a cosine similarity threshold of 0.95, with at most one removal per second during dead-time. \mname{} processes the video stream at 2 fps. All offline and streaming inference experiments are conducted on a single H100 GPU.

\begin{table}[t]
\centering
\setlength{\tabcolsep}{3.5mm}{
\caption{\textbf{Sensitivity to occupancy budget $K$ on Detection tasks.} Each cell reports Timeliness-F1 (\%), with gains from $K=1$ to $K=5$ in \textcolor{green!50!black}{green}.}
\label{tab:supp:K_sensitivity}
\resizebox{\linewidth}{!}{
\begin{tabular}{@{}l c c c ccc ccc@{}}
\toprule
\textbf{Model} & \textbf{\#} & \textbf{Post} & \textbf{Model} &
\multicolumn{3}{c}{\textbf{ABD}} &
\multicolumn{3}{c}{\textbf{PNR}} \\
\cmidrule(lr){5-7}\cmidrule(lr){8-10}
& \textbf{Frames} & \textbf{Train?} & \textbf{Size} &
$K=1$ & $K=5$ & \textbf{Gain} &
$K=1$ & $K=5$ & \textbf{Gain} \\
\midrule

\rowcolor{headergray}
\multicolumn{10}{l}{\textbf{Offline MLLM}} \\
\; Qwen2.5-VL (8 frames)~\cite{bai2025qwen2}
& 8 & \xmark & 7B
& 4.5 & 10.0 & \textcolor{green!50!black}{+5.5}
& 4.3 & 8.1 & \textcolor{green!50!black}{+3.8} \\
\midrule

\rowcolor{headergreen}
\multicolumn{10}{l}{\textbf{Streaming VideoLLMs}} \\
\; StreamingVLM~\cite{xu2025streamingvlm}
& 2fps & \cmark & 7B
& 2.7 & 3.9 & \textcolor{green!50!black}{+1.2}
& 1.3 & 2.5 & \textcolor{green!50!black}{+1.2} \\
\; MMDuet2~\cite{wang2025mmduet2}
& & & & & & & & & \\
\qquad + KVflush
& 2fps & \cmark & 3B
& 11.7 & 14.2 & \textcolor{green!50!black}{+2.5}
& 4.3 & 5.1 & \textcolor{green!50!black}{+0.8} \\
\qquad + StreamingVLM
& 2fps & \cmark & 3B
& 11.7 & 13.8 & \textcolor{green!50!black}{+2.1}
& 1.3 & 1.6 & \textcolor{green!50!black}{+0.3} \\
\; MiniCPM-o 4.5~\cite{cui2026minicpm}
& 2fps & \cmark & 9B
& 14.9 & 15.4 & \textcolor{green!50!black}{+0.5}
& 14.1 & 14.3 & \textcolor{green!50!black}{+0.2} \\
\; JoyAI-VL-Interaction~\cite{yao2026joyai}
& 2fps & \cmark & 8B+4B
& 16.6 & 17.0 & \textcolor{green!50!black}{+0.4}
& 14.7 & 15.6 & \textcolor{green!50!black}{+0.9} \\

\addlinespace[0.6ex]
\arrayrulecolor{gray!80}\cmidrule(lr){1-10}\arrayrulecolor{black}
\; \textbf{\mname{}} & & & & & & & & & \\
\qquad w/ Qwen2.5-VL
& 2fps & \xmark & 7B
& 20.8 & 22.6 & \textcolor{green!50!black}{+1.8}
& 17.9 & 19.5 & \textcolor{green!50!black}{+1.6} \\
\bottomrule
\end{tabular}}}
\vspace{-0.3cm}
\end{table}

\begin{figure}[t]
\centering
\vspace{0.3cm}
\begin{minipage}[t]{0.47\linewidth}
    \centering
    \includegraphics[width=\linewidth]{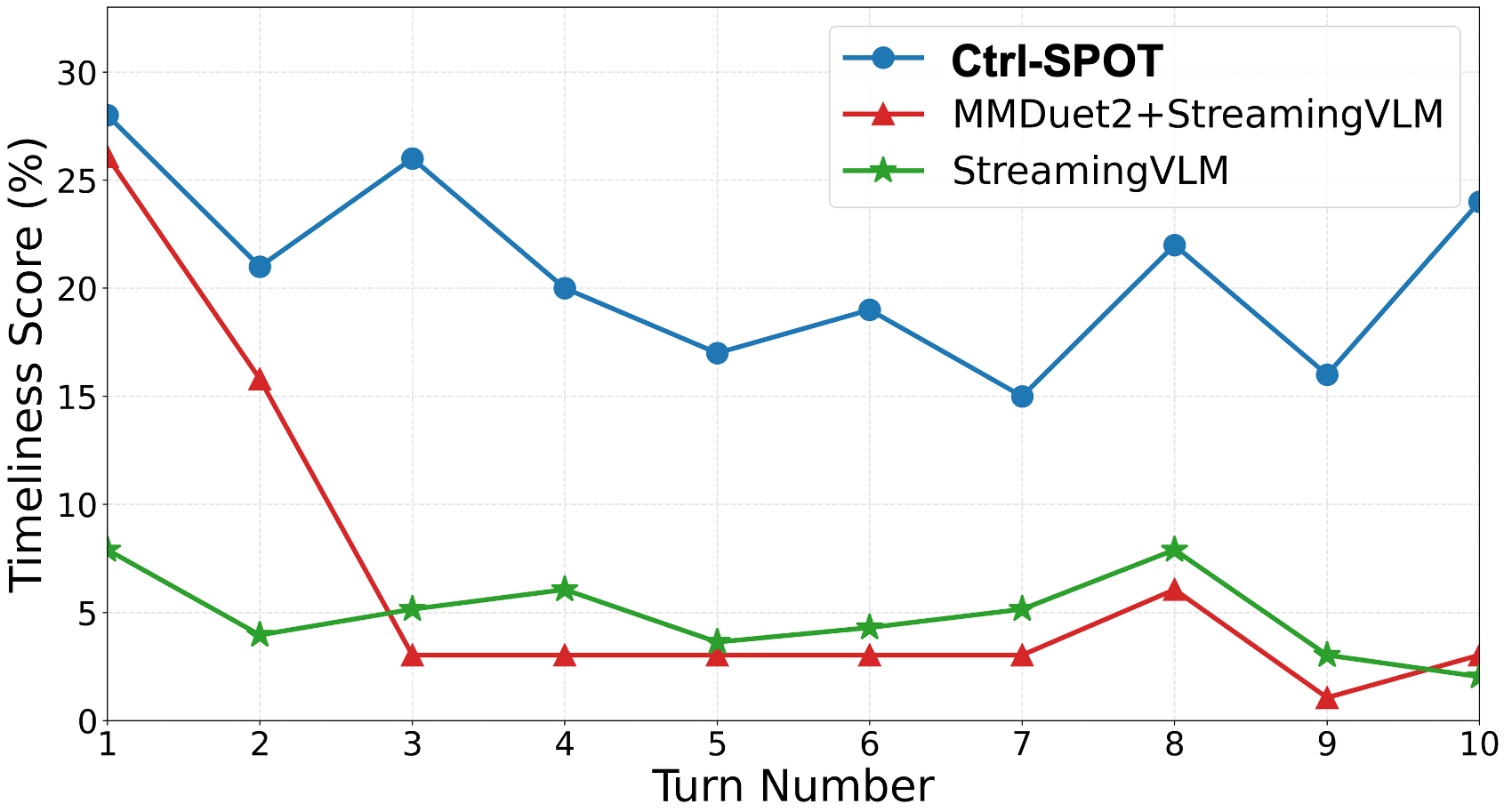}
    \caption{\textbf{Performance across multiple turns} on a subset of PNR videos.}
    \label{fig:multiturn}
\end{minipage}
\hfill
\begin{minipage}[t]{0.49\linewidth}
    \centering
    \includegraphics[width=\linewidth]{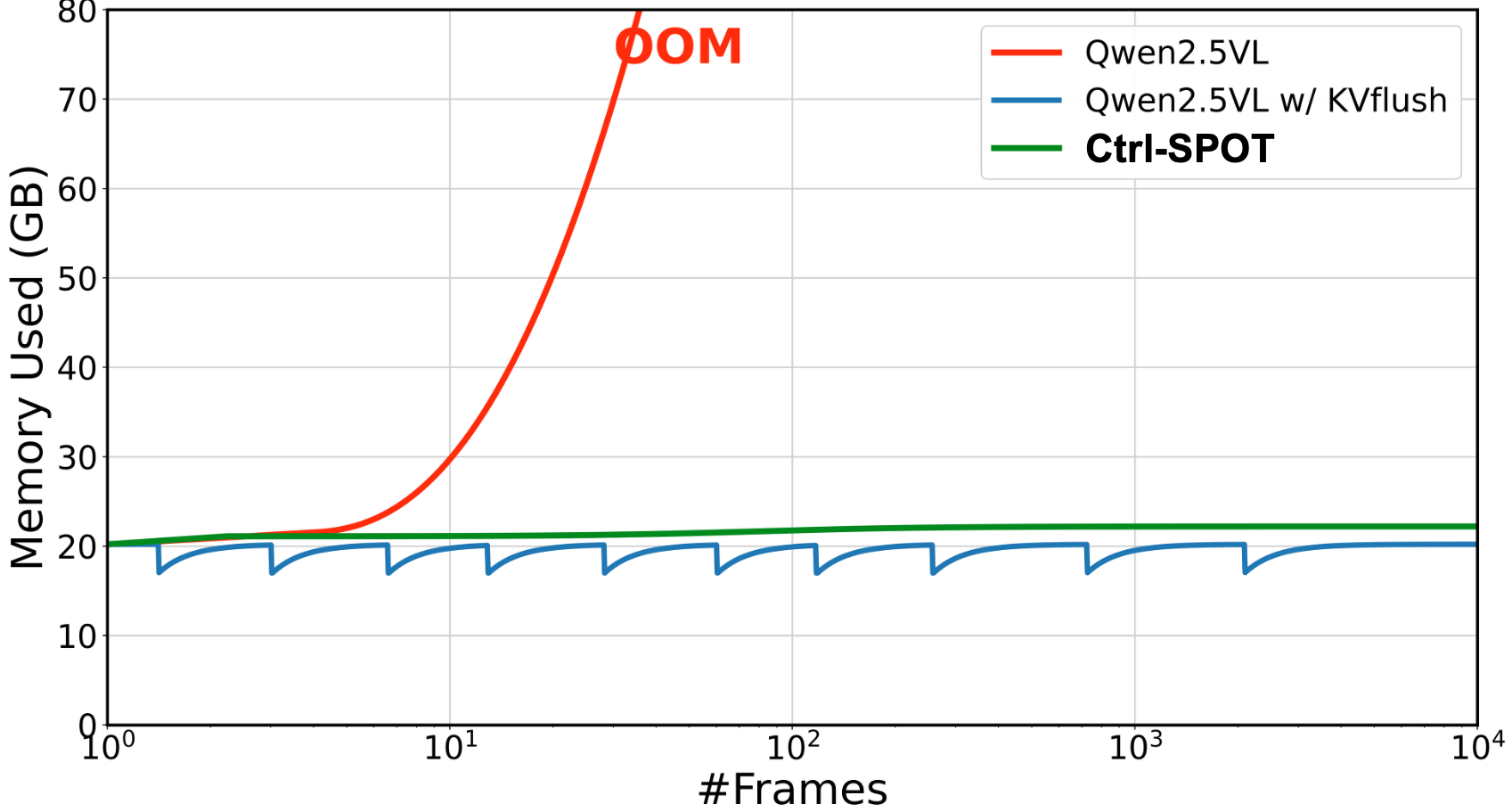}
    \caption{\textbf{Memory usage analysis} on one H100 GPU. OOM refers to out of memory.}
    \label{fig:memory_analysis}
\end{minipage}
\vspace{-0.5cm}
\end{figure}

\vspace{-0.2cm}

\subsection{Ablations \& Analysis}\label{app:ablations}

\noindent\textbf{Sensitivity to $K$ reveals response redundancy.}
As shown in~\cref{tab:supp:K_sensitivity}, increasing $K$ from 1 to 5 substantially improves T-F1 for naively prompted Qwen2.5-VL, consistent with repeated predictions (response echoing) across nearby frames. StreamingVLM's performance is too low to benefit meaningfully from increasing $K$, despite frequent over-responses, as semantically invalid predictions remain FPs regardless of $K$. In contrast, post-trained models like MMDuet2, and especially interaction models like MiniCPM-o 4.5 and JoyAI-VL-Interaction, exhibit minimal gains, suggesting that post-training encourages more selective responses. Moreover, under-responsiveness (FNs) cannot be mitigated by increasing $K$. \mname{} shows moderate T-F1 gains, but substantially smaller relative gains than naive prompting, demonstrating that the controller effectively mitigates over-response while preserving strong performance.

\noindent\textbf{Why do open-source models fail on Interaction and Intervention?}
Despite high human annotator agreement, especially for Intervention (\cref{fig:supp::intervals}), and strong closed-source performance, most open-source models collapse on these tasks~(\cref{tab:offlineevals}). GPT-5 and Gemini-3-Flash score consistently well, indicating that the tasks are well-defined and solvable. The failures are model-specific: \\
\textit{(i)} For SQA, most failures stem from incorrect responses: models ignore the question and describe arbitrary events in the scene. \\
\textit{(ii)} SPG exposes a fundamental weakness. Most open-source models score 0\%, while GPT-5 and Gemini-3-Flash reach roughly 30\%. SPG requires the model to continuously understand a long video as the user is performing their task, observe actions, monitor task progress, and then proactively issue the next step guidance. Only the closed-source models reliably recall and follow the action instructions over long contexts. \\
\textit{(iii)} SI follows the same pattern but is harder, as the model now needs to reason whether the user is struggling and issue an appropriate intervention. \\
\textit{(iv)} UI is the hardest case. There is no explicit query; the model must detect subtle cues and intervene at the correct moment. Closed-source models perform reasonably well, with GPT-5 scoring over 50\%. InternVL-3.5 handles some cases, but most open-source models fail entirely. Notably, InternVL-3.5 often produces responses continuously; its outputs are generally valid interventions or guidance, and a few happen to coincide with the correct moment, yielding non-zero scores.

\noindent\textbf{Timeliness-score across multiple turns.} \cref{fig:multiturn} shows timeliness across increasing turn numbers on a subset of PNR videos with $>$10 turns. StreamingVLM remains low but stable due to its live-commentary design producing frequent query-agnostic responses. MMDuet2 with StreamingVLM’s KV strategy degrades quickly as turns increase, likely because it was not trained to handle multiple active queries. In contrast, \mname{} maintains stable performance across turns, demonstrating the benefit of persistent query state.

\noindent\textbf{Memory analysis.} \cref{fig:memory_analysis} shows GPU memory usage during streaming. Qwen2.5-VL out-of-the-box grows rapidly with video length, running out-of-memory quickly. KVflush maintains bounded memory by maintaining a sliding window and clearing the cache at query onset. \mname{} keeps memory bounded while preserving relevant context through its long-short term KV design.

\begin{figure}[t]
\centering

\begin{minipage}[t]{0.58\textwidth}
\centering
\includegraphics[width=\linewidth]{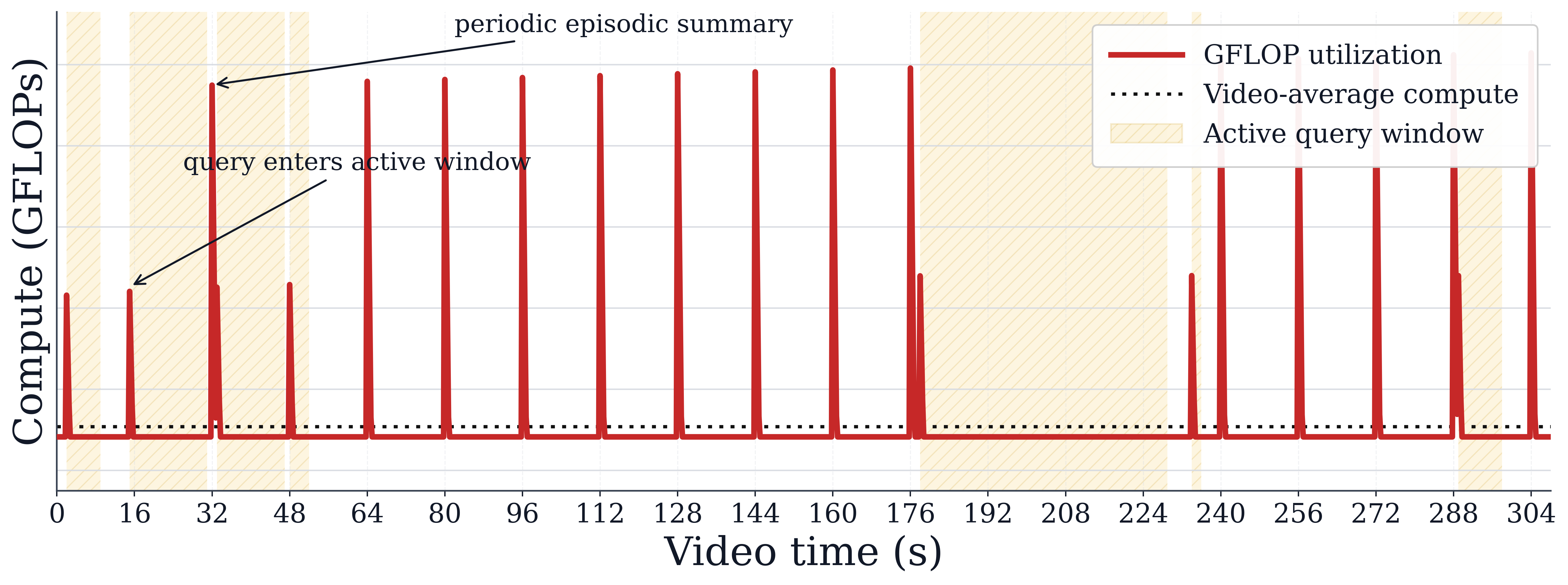}
\caption{\textbf{Compute utilization} during dead vs. active times on a 5 minute PNR video on 1 H100.}
\label{fig:supp::deadcompute}
\end{minipage}
\hfill
\begin{minipage}[t]{0.38\textwidth}
\centering
\includegraphics[width=\linewidth]{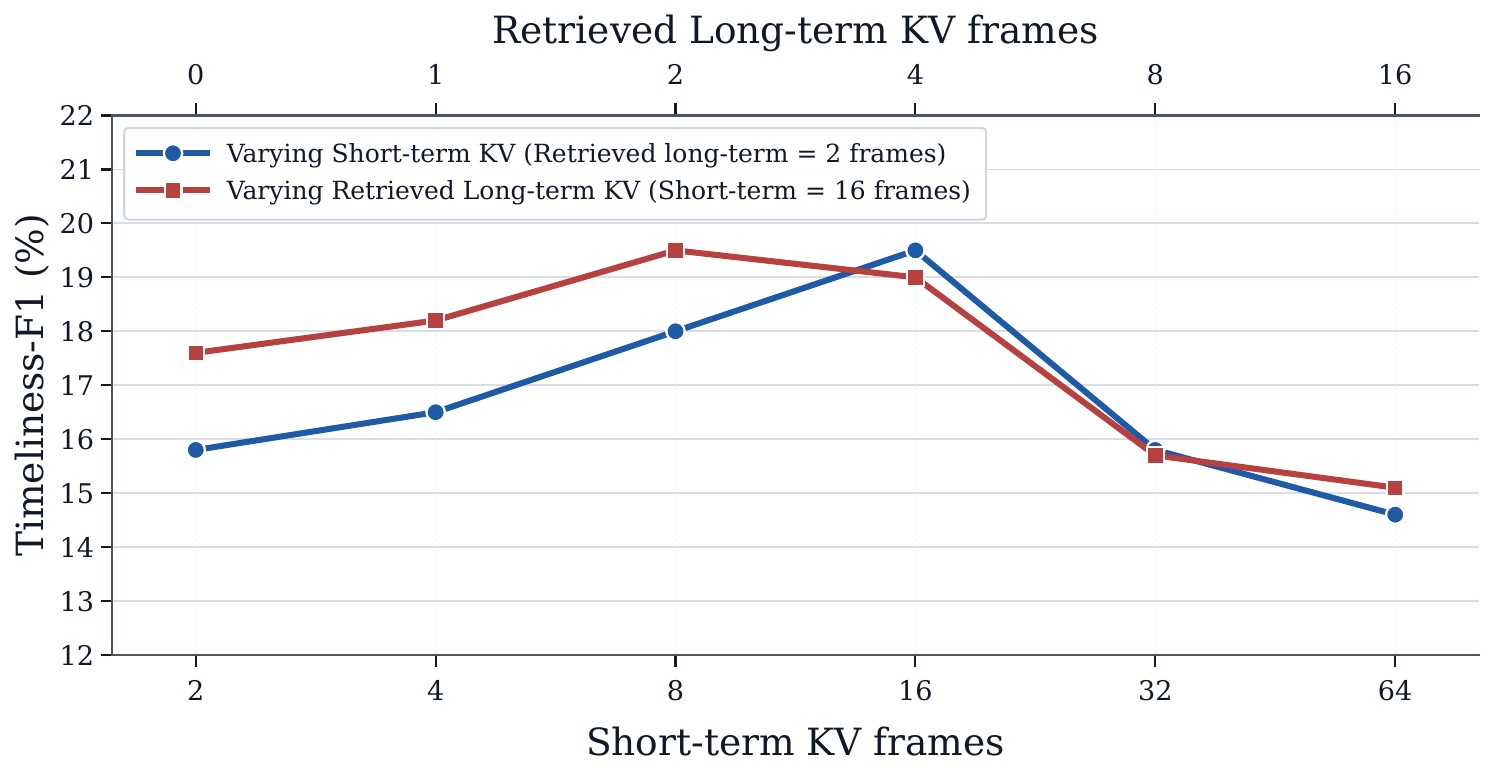}
\caption{\textbf{Ablating short vs long-term KV} on PNR.}
\label{fig:supp::kvsplit}
\end{minipage}

\vspace{-0.3cm}
\end{figure}

\noindent\textbf{Compute utilization.}
To understand how \mname{} allocates computation over time, we plot GFLOPs per second in~\cref{fig:supp::deadcompute}. Two patterns emerge. First, we do not zero-out compute during dead-time. The model continues lightweight streaming updates to maintain short-term state and remain responsive. Second, dead-time enables additional computation: episodic summarization runs periodically and produces the recurring spikes in utilization. When a query arrives, compute is spent retrieving relevant frames from CPU to GPU and performing a forward pass to prepare the model for response generation. As a result, \mname{} spreads computation across the entire stream rather than concentrating it at response time. This reflects the core idea of the method: dead-time can be used to scale compute without increasing response latency.

\noindent\textbf{Ablating long-short term KV.}
\cref{fig:supp::kvsplit} visualizes the effect of short- and long-term KVs on the PNR task measured by Timeliness-F1. Short-term KV has the larger impact. Increasing $N_S$ from 2 to 16 frames increases T-F1, showing that recent context is critical for detecting the precise moment of an event. Long-term retrieval provides a smaller but complementary gain. Performance peaks at $K{=}2$ retrieved frames, remains stable till 4, and declines afterwards, suggesting that too many distant frames introduce irrelevant context. Notably, even with $K{=}0$, the red curve starts at 17.6\%, confirming that short-term memory carries most of the perceptual signal. We therefore select $N_S{=}16$ and $N_L{=}256$ (top-$K{=}2$ retrieved per query) as the default, balancing T-F1 and memory usage.

\section{Task Exemplars}\label{app:exemplars}
\begin{figure*}[h]
\centering
\includegraphics[width=0.8\linewidth]{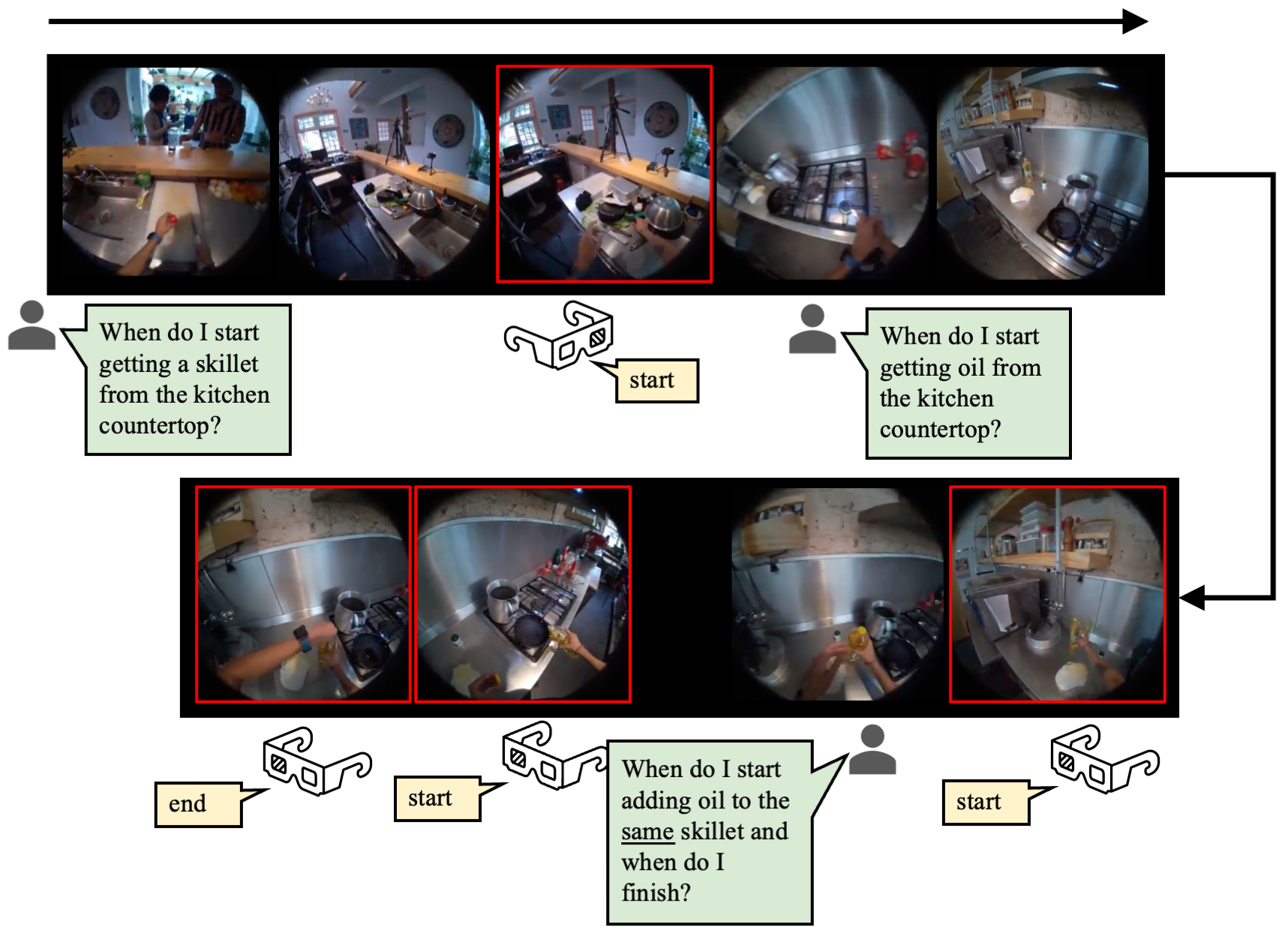}
\caption{\textbf{An illustrated example of the Action Boundary Detection (ABD) task.} The highlighted ``\underline{same}'' is \textit{referential},~\ie~it points back to the skillet mentioned in the first query.}
\label{fig:supp::abd}
\end{figure*}

\begin{figure*}[h]
\centering
\includegraphics[width=\linewidth]{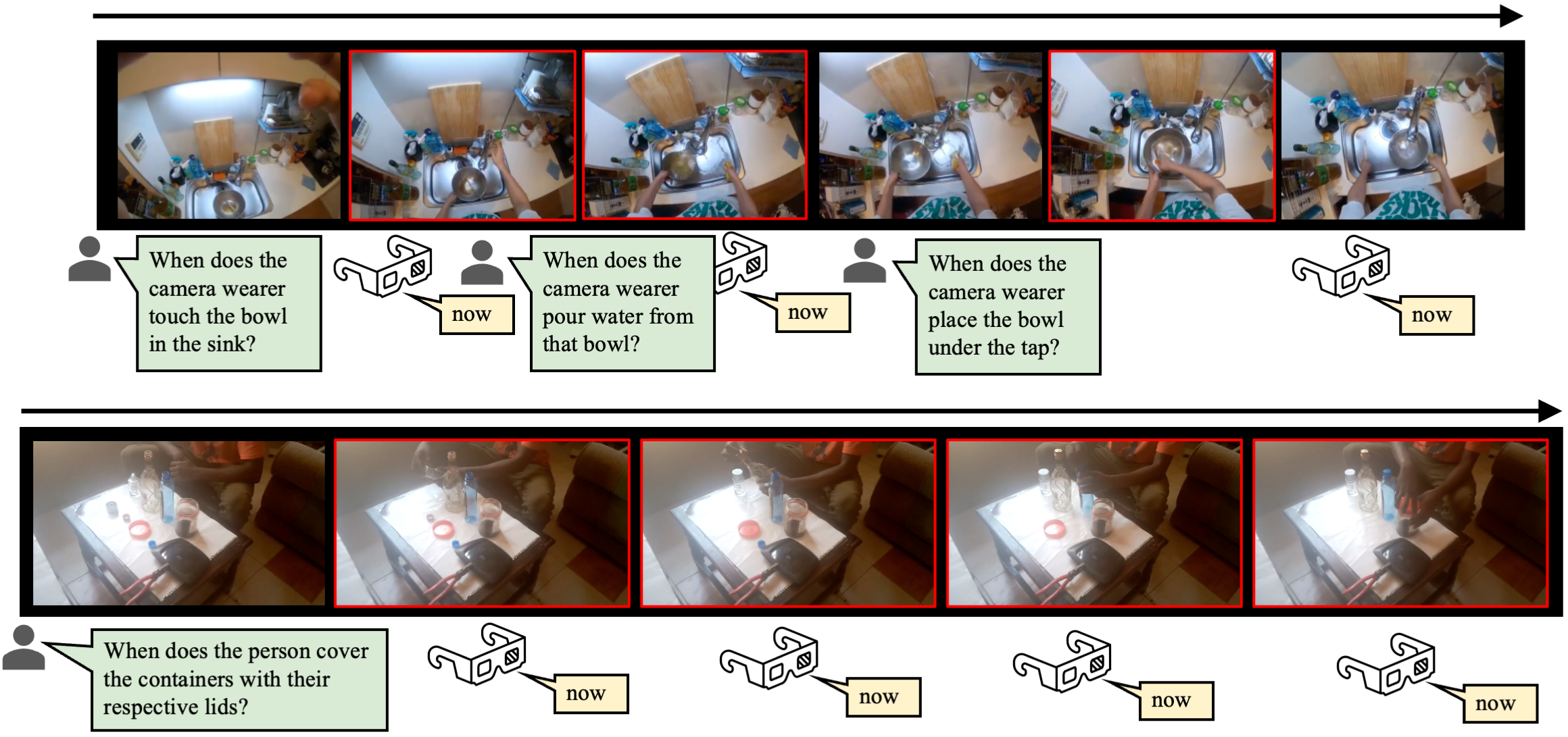}
\caption{\textbf{Illustrated examples of the Point-of-No-Return Detection (PNR) task.} The top example shows a multi-turn, single-response case; the bottom example shows a single-turn, multi-response case.}
\label{fig:supp::pnr}
\end{figure*}

\begin{figure*}[h]
\centering
\includegraphics[width=\linewidth]{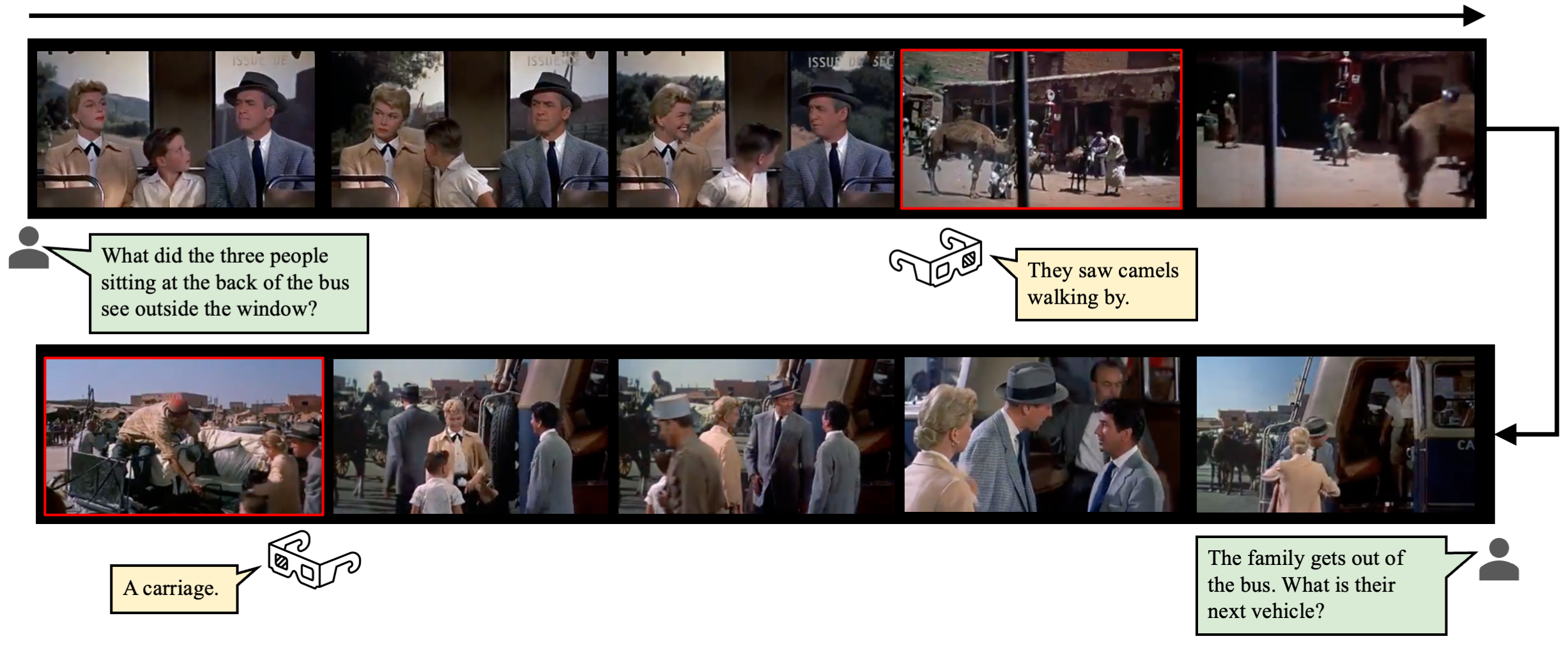}
\caption{\textbf{An illustrated example of the Streaming Question Answering (SQA) task.} The gap between ask time and response time is typically larger for this task.}
\label{fig:supp::sqa}
\end{figure*}

\begin{figure*}[h]
\centering
\includegraphics[width=\linewidth]{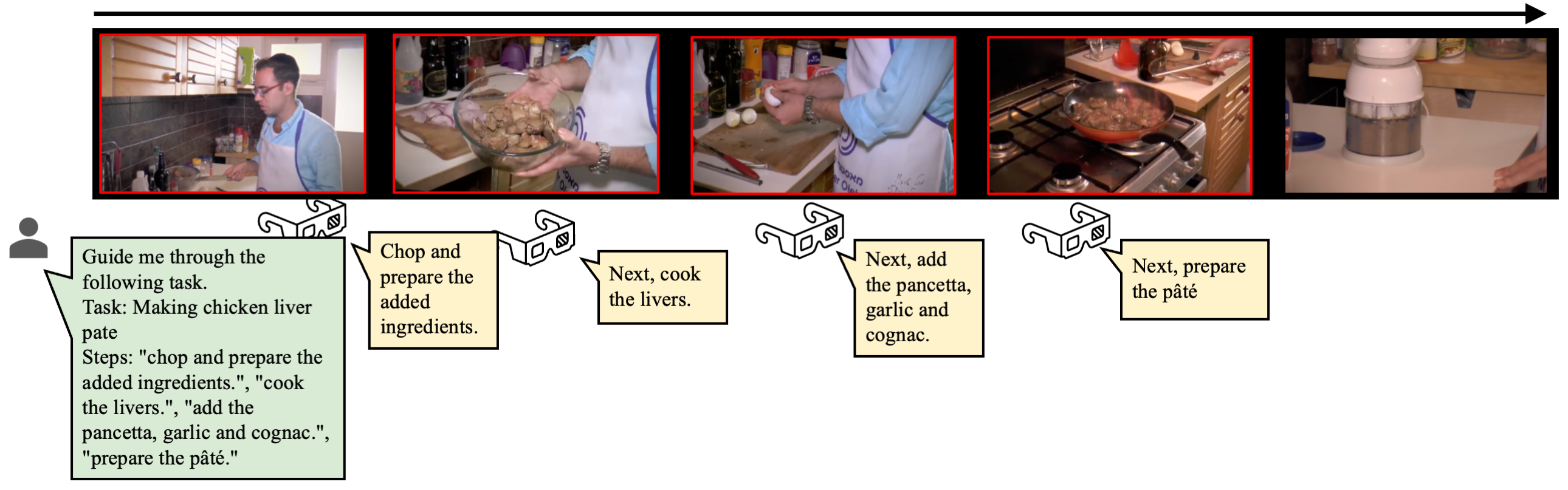}
\caption{\textbf{An illustrated example of the Streaming Procedural Guidance (SPG) task.} 
A set of allowed action labels is given upfront, and the model must respond with the correct next action at the correct times as the video streams.}
\label{fig:supp::spg}
\end{figure*}

\begin{figure*}[h]
\centering
\includegraphics[width=\linewidth]{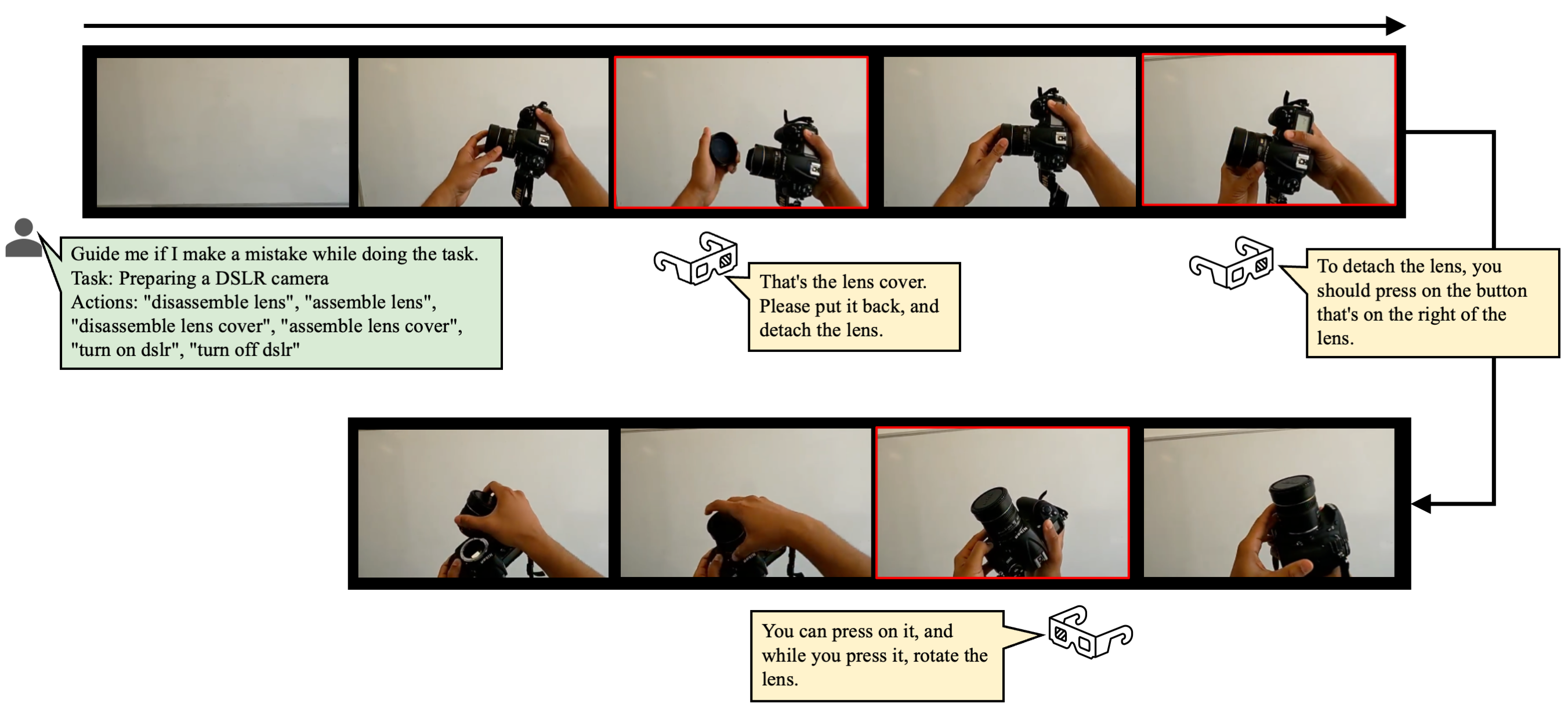}
\caption{\textbf{An illustrated example of the Solicited Intervention (SI) task.} The user provides a task description and a set of allowed actions, and the model issues corrective guidance only when it detects an error in the streaming video.}
\label{fig:supp::si}
\end{figure*}

\begin{figure*}[h]
\centering
\includegraphics[width=0.8\linewidth]{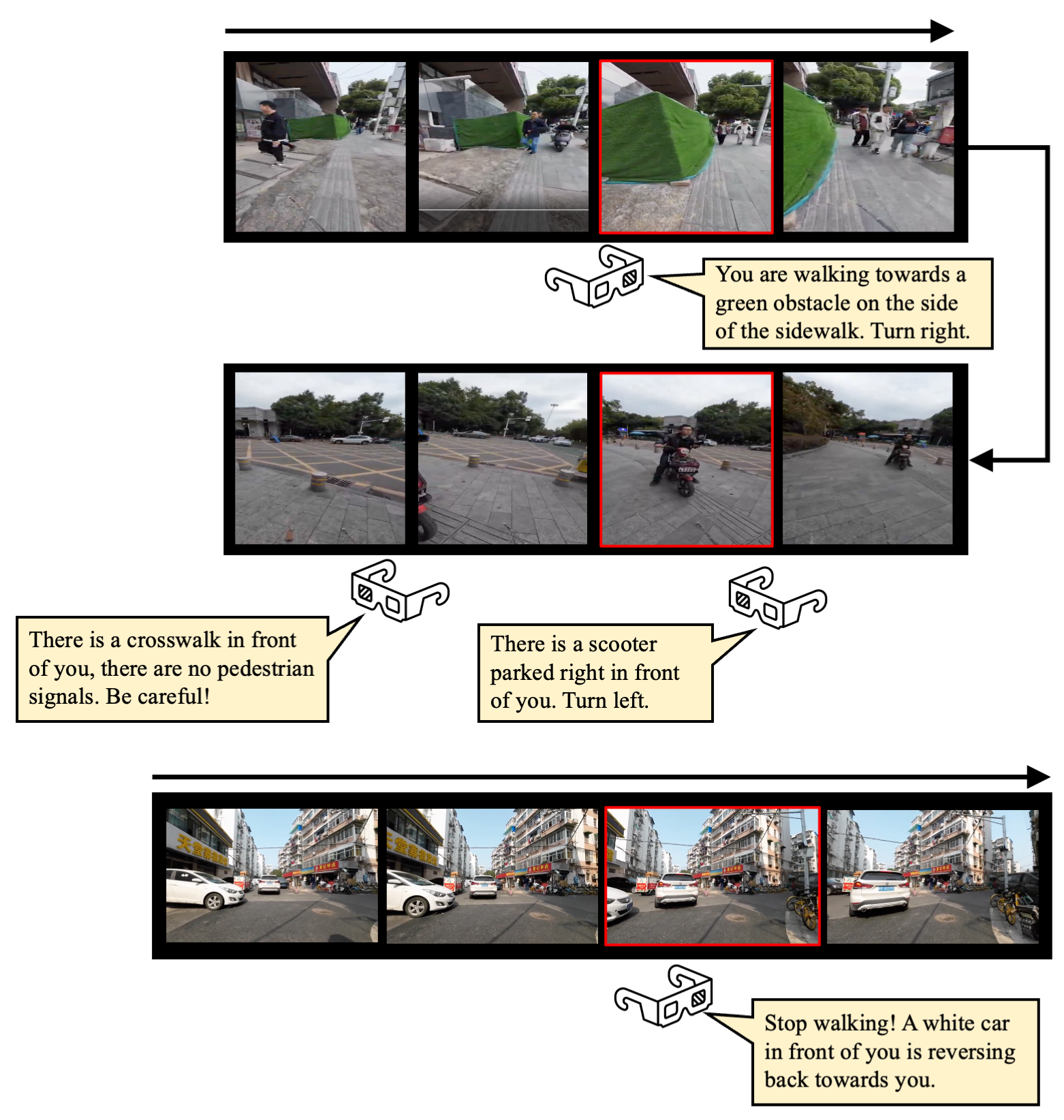}
\caption{\textbf{Illustrated examples of the Unsolicited Intervention (UI) task.} 
The top example shows a multi-response scenario, where the model issues several warnings as new hazards appear; the bottom example shows a single-response case.}
\label{fig:supp::ui}
\end{figure*}

We include representative exemplars for each SPOT-Bench task: ABD (Fig.~\ref{fig:supp::abd}), PNR (Fig.~\ref{fig:supp::pnr}), SQA (Fig.~\ref{fig:supp::sqa}), SPG (Fig.~\ref{fig:supp::spg}), SI (Fig.~\ref{fig:supp::si}), and UI (Fig.~\ref{fig:supp::ui}).

\begin{figure*}[h]
    \begin{promptbox}[prompt:ABD]{Prompt for ABD keyframe generation}
    Choose all applicable boundary types for the step. Reply with EXACTLY ONE of:\\
    start | end | start,end | none
    
    \vspace{0.5em}
    Meanings (short \& strict)
    \begin{itemize}[leftmargin=*, noitemsep, topsep=2pt, label=-]
        \item start = earliest visually grounded onset unique to this step (first intentional motion that commits to it---grasp/tear/approach specific to the step).
        \item end = the moment the step is achieved/committed (state flip or completion---open/closed, removed/inserted, switched on/off, fully assembled).
        \item start,end = both boundaries are reasonably observable in typical egocentric video; prefer this when in doubt and both are defensible.
        \item none = use sparingly; step is too vague/ambient to anchor precise boundaries
    \end{itemize}
    
    \vspace{0.5em}
    Decision rules:
    \begin{itemize}[leftmargin=*, noitemsep, topsep=2pt, label=-]
        \item If a discrete state flip completes the step $\rightarrow$ include end.
        \item If there is a clear initiating motion unique to this step $\rightarrow$ include start.
        \item If both are observable with normal egocentric cues $\rightarrow$ choose start,end.
        \item Use none only when neither boundary can be pinned down reliably.
    \end{itemize}
    
    \vspace{0.5em}
    Few-shot Examples:
    ...
    
    \vspace{0.5em}
    Now answer for this instance (ONE TOKEN ONLY from the allowed set):\\
    step\_name: \{step\_name\}\\
    step\_description: \{step\_description\}
    \end{promptbox}
\end{figure*}

\begin{figure*}[h]
    \begin{promptbox}[prompt:PNR]{Prompt for PNR keyframe generation}
    Choose the single best key-frame type.
    
    Reply with EXACTLY ONE lowercase word (no extra text):\\
    contact | pnr | impossible
    
    \vspace{0.5em}
    Meanings (short \& strict)
    \begin{itemize}[leftmargin=*, noitemsep, topsep=2pt, label=-]
        \item contact = the first relevant physical touch for the described action (e.g. pick, grab, press, tap, hold, adjust, scrape, stir, wipe).
        \item pnr = the key moment is defined by a discrete state flip or commit (e.g. open/close, insert/remove, attach/detach, lock/unlock, turn on/off).
        \item impossible = use sparingly; only when the narration clearly involves neither subject/object contact nor a state change.
    \end{itemize}
    
    \vspace{0.5em}
    Decision rules:
    \begin{itemize}[leftmargin=*, noitemsep, topsep=2pt, label=-]
        \item If the action produces a clear state flip $\rightarrow$ pnr.
        \item Else if the action involves physical contact that grounds the narration $\rightarrow$ contact.
        \item Else $\rightarrow$ impossible (rare).
    \end{itemize}
    
    Use hints from both verb and state\_transition if present.
    
    \vspace{0.5em}
    Few-Shot Examples:
    ...
    
    \vspace{0.5em}
    Now answer for this instance (ONE WORD ONLY):\\
    narration: \{narration\}\\
    verb: \{verb\}\\
    state\_transition: \{state\_transition\}
    \end{promptbox}
\end{figure*}

% --- SQA PROMPT ---
\begin{figure*}[t!]
    \begin{promptbox}[prompt:SQA]{SQA: GPT-as-Judge prompt for matching}
You are an automatic evaluator for video question answering.
Your job is to judge whether a model's predicted answer meaningfully matches the
correct answer for a video-based question.

INSTRUCTIONS:
\begin{itemize}[leftmargin=*, noitemsep, topsep=2pt, label=-]
    \item Focus strictly on semantic meaning.
    \item Paraphrases are fine if they preserve meaning.
    \item If the predicted answer does NOT answer the question, mark "no".
    \item If it answers the same underlying fact, mark "yes".
    \item Score meaning similarity from 0-5:
      5 = same meaning
      4 = near-identical
      3 = partially correct but acceptable
      2 = weakly related
      1 = barely related
      0 = unrelated or wrong
    \item No explanations. Only output JSON.
\end{itemize}

Evaluate:\\
Question: \{question\}\\
Correct Answer: \{gold\}\\
Predicted Answer: \{pred\}

Respond ONLY with:\\
\{"pred": "yes" or "no", "score": INTEGER\}
    \end{promptbox}
\end{figure*}

% --- SPG PROMPT ---
\begin{figure*}[t!]
    \begin{promptbox}[prompt:SPG]{SPG: GPT-as-Judge prompt for matching}
You are an automatic evaluator for procedural guidance.
Your job is to check whether a model's predicted next action meaningfully matches
the correct next action for a video-based procedural task.

INSTRUCTIONS:
\begin{itemize}[leftmargin=*, noitemsep, topsep=2pt, label=-]
    \item Compare predicted action vs correct action purely on meaning.
    \item Accept paraphrases that represent the same action or intent.
    \item If the predicted action does NOT correspond to the same step, mark "no".
    \item Score similarity on a 0-5 scale:
      5 = identical step
      4 = equivalent phrasing
      3 = partial correctness
      2 = weak overlap
      1 = barely related
      0 = wrong action
    \item No reasoning or extra text.
\end{itemize}

Evaluate:\\
Correct Action: \{gold\}\\
Predicted Action: \{pred\}

Respond ONLY with:\\
\{"pred": "yes" or "no", "score": INTEGER\}
    \end{promptbox}
\end{figure*}

% --- SI PROMPT ---
\begin{figure*}[t!]
    \begin{promptbox}[prompt:SI]{SI: GPT-as-Judge prompt for matching}
You are an automatic evaluator for an intervention task.
Here the assistant gives a short instruction to help a user with mistakes or hesitation.

Your job is to judge whether the model's predicted instruction meaningfully
matches the correct intended instruction.

INSTRUCTIONS:
\begin{itemize}[leftmargin=*, noitemsep, topsep=2pt, label=-]
    \item Focus only on meaning: does the instruction guide the user toward the same next step or same correction.
    \item Paraphrasing is allowed if the guidance is equivalent.
    \item If the predicted instruction does not help in the same way, mark "no".
    \item Score similarity from 0-5:
      5 = identical guidance
      4 = equivalent instruction
      3 = partially correct but usable
      2 = weak help but still helps the user
      1 = almost irrelevant
      0 = wrong instruction
    \item No explanation text.
\end{itemize}

Evaluate:\\
Correct Instruction: \{gold\}\\
Predicted Instruction: \{pred\}

Respond ONLY with:\\
\{"pred": "yes" or "no", "score": INTEGER\}
    \end{promptbox}
\end{figure*}

% --- UI PROMPT ---
\begin{figure*}[t!]
    \begin{promptbox}[prompt:UI]{UI: GPT-as-Judge prompt for matching}
You are an automatic evaluator for an blind assistance intervention task,
where an assistant gives short spoken warnings or guidance based on a blind
user's video.

Your job is to judge whether the model's predicted warning or guidance meaningfully
matches the correct warning based on the visible risk.

INSTRUCTIONS:
\begin{itemize}[leftmargin=*, noitemsep, topsep=2pt, label=-]
    \item Judge semantic meaning only.
    \item A correct prediction must warn about the same risk or obstacle.
    \item Paraphrases are fine; irrelevant or mismatched warnings are incorrect.
    \item Score similarity on a 0-5 scale:
      5 = same warning
      4 = near-identical
      3 = partially correct
      2 = weak relevance
      1 = barely related but still a warning
      0 = wrong/no warning
    \item No explanations.
    \item Sometimes multiple correct warnings are provided separated by semicolons;
\end{itemize}

Evaluate:\\
Correct Warning separated by ',' or ';': \{gold\}\\
Predicted Warnings: \{pred\}

Respond ONLY with:\\
\{"pred": "yes" or "no", "score": INTEGER\}
    \end{promptbox}
\end{figure*}

\end{document}